%% file: main.tex
\pdfoutput=1
\documentclass[10pt,letterpaper]{article}

\usepackage{glancenav}
\input{macros}

\usepackage[colorlinks,breaklinks,linkcolor=linkblue,citecolor=linkblue,urlcolor=accentdark]{hyperref}
\usepackage[capitalize,nameinlink]{cleveref}
\crefname{section}{Sec.}{Secs.}
\Crefname{section}{Section}{Sections}
\crefname{table}{Tab.}{Tabs.}
\Crefname{table}{Table}{Tables}
\crefname{figure}{Fig.}{Figs.}
\Crefname{figure}{Figure}{Figures}
\crefname{equation}{Eq.}{Eqs.}
\crefname{algorithm}{Alg.}{Algs.}
\Crefname{algorithm}{Algorithm}{Algorithms}
\crefname{appendix}{App.}{Apps.}
\Crefname{appendix}{Appendix}{Appendices}

\input{authors}
\hypersetup{
  pdftitle={GlanceNav: Efficient Vision-and-Language Navigation with Streaming Memory and Counterfactual Rewind}}
\hypersetup{pdfauthor={Anh Dao, Quan-Dung Pham, Danh Vinh Le, Tuyen Pham Le, Quan Nguyen, Yuzhang Shang}}

\begin{document}

\makepaperheader

\begin{abstractbox}
\input{sec/0_abstract}
\end{abstractbox}

\begin{figure}[!b]
\centering
\input{fig/teaser}
\end{figure}

\input{sec/1_intro}
\input{sec/2_related}
\input{sec/3_method}
\input{sec/4_experiments}
\input{sec/5_conclusion}

{\small
\setlength{\bibsep}{5pt plus 3pt minus 1pt}
\bibliographystyle{ieeenat_fullname}
\bibliography{main}
}

\clearpage
\appendix
\input{sec/A_appendix}

\end{document}

%% file: macros.tex
\usepackage{algorithm}
\usepackage{algpseudocode}
\usepackage{tikz}
\usetikzlibrary{arrows.meta,positioning,calc}
\usepackage{pgfplots}
\pgfplotsset{compat=1.18}

\definecolor{ours}{RGB}{230,120,20}
\definecolor{oursbg}{RGB}{253,238,222}
\definecolor{stateC}{RGB}{108,84,168}
\definecolor{winC}{RGB}{52,120,176}
\definecolor{zeroC}{RGB}{0,140,140}
\definecolor{extC}{RGB}{130,130,130}
\definecolor{inkC}{RGB}{60,60,60}
\definecolor{refgray}{RGB}{128,128,128}

\tikzset{fig/.style={font=\sffamily\scriptsize, line join=round, line cap=round}}

\newcommand{\method}{GlanceNav\xspace}
\newcommand{\core}{CoRe\xspace}
\newcommand{\cmark}{\ding{51}}
\newcommand{\xmark}{\ding{55}}
\DeclareMathOperator*{\argmin}{arg\,min}
\newcommand{\norm}[1]{\left\lVert #1 \right\rVert}
\newcommand{\R}{\mathbb{R}}
\newcommand{\vis}{\mathcal{V}}
\newcommand{\bnk}{\mathcal{B}}
\newcommand{\grp}{\mathcal{G}}
\newcommand{\pre}{\mathcal{P}}

\newcommand{\para}[1]{\par\addvspace{3.5pt plus 1.5pt minus 1pt}\noindent\textbf{#1}\hspace{0.45em}}

\newcolumntype{+}{>{\global\let\currentrowstyle\relax}}
\newcolumntype{^}{>{\currentrowstyle}}
\newcommand{\rowstyle}[1]{\gdef\currentrowstyle{#1}#1\ignorespaces}
\newcommand{\rf}{\rowstyle{\color{refgray}}}

\algrenewcommand\algorithmiccomment[1]{\hfill{\footnotesize\textcolor{extC}{$\triangleright$ #1}}}

\makeatletter
\newcommand{\tabpanel}[2][0.485]{%
  \begin{minipage}[t]{#1\linewidth}\vspace{0pt}\def\@captype{table}\centering\input{#2}\end{minipage}}
\newcommand{\figpanel}[2][0.485]{%
  \begin{minipage}[t]{#1\linewidth}\vspace{0pt}\def\@captype{figure}\centering\input{#2}\end{minipage}}
\makeatother
\newlength{\rwimgw}
\newcommand{\rwframe}[1]{{\setlength{\fboxsep}{0pt}\setlength{\fboxrule}{0.35pt}%
  \fcolorbox{black!22}{white}{\adjustbox{height=1.7763\rwimgw,Clip=0pt 0pt {\width-\rwimgw} 0pt}{\includegraphics{#1}}}}}
\newcommand{\rwrow}[7][2]{%
  \setlength{\rwimgw}{\dimexpr 0.232\linewidth-0.7pt\relax}%
  \parbox[b][#1\dimexpr 8.4pt\relax][t]{\linewidth}{\raggedright\sffamily\scriptsize\color{inkC}%
    \textbf{#2}\enspace\textit{``#3''}\strut}\par\nointerlineskip\vspace{2pt}%
  \noindent\rwframe{#4}\hfill\rwframe{#5}\hfill\rwframe{#6}\hfill\rwframe{#7}\par}
\newcommand{\rwpanel}[7][2]{\begin{minipage}[t]{0.485\linewidth}\rwrow[#1]{#2}{#3}{#4}{#5}{#6}{#7}\end{minipage}}
\newenvironment{sidecap}[1][0.40]{%
  \begin{minipage}[c]{#1\linewidth}\captionsetup{skip=0pt}}{\end{minipage}\hfill}
\newenvironment{sidebody}[1][0.56]{%
  \begin{minipage}[c]{#1\linewidth}\centering}{\end{minipage}}

%% file: authors.tex
\papertitle{{\color{accent}GlanceNav}: Efficient Vision-and-Language Navigation with\\[1pt] Streaming Memory and Counterfactual Rewind}
\runningtitle{GlanceNav: Efficient Vision-and-Language Navigation with Streaming Memory and Counterfactual Rewind}

\paperauthors{%
  \pauthor{Anh Dao}{1,2}\quad
  \pauthor{Quan-Dung Pham}{2}\quad
  \pauthor{Danh Vinh Le}{2}\quad
  \pauthor{Tuyen Pham Le}{2}\quad
  \pauthor{Quan Nguyen}{2,3}\quad
  \pauthor{Yuzhang Shang}{1}}

\paperaffiliations{%
  \textsuperscript{1}University of Central Florida\qquad
  \textsuperscript{2}VinMotion\qquad
  \textsuperscript{3}University of Southern California}

\projecturl{https://glancenav.github.io/}
\codeurl{https://github.com/GlanceNav/GlanceNav}
\modelurl{}

%% file: sec/0_abstract.tex
Multimodal large language models (MLLMs) have become strong policies for vision-and-language navigation (VLN). However, most of them encode a sample of past frames again at every decision and are trained on isolated decisions, so computation is redundant and the usable history is short. They also learn from their own mistakes through outcome rewards that give one scalar to a whole episode. We ask whether a navigator can encode each frame only once, at a constant cost per step, and still remember what matters and learn efficiently from its mistakes. We present GlanceNav, a streaming navigator built on a hybrid MLLM. Its Gated DeltaNet layers carry a recurrent state through the whole episode, while its attention layers keep a short window of recent observations and a small bank of keyframes, selected by how much each frame changes the recurrent state. This constant-size memory keeps the cost per step fixed and allows training on whole episodes with the same memory as at inference. Because the entire memory can also be copied in constant time, we further propose Counterfactual Rewind (CoRe), a reinforcement learning stage that rewinds a drifting rollout to its closest approach to the goal, lets an expert correct a few actions, and credits decisions by the resulting change in outcome. Trained only on the R2R-CE and RxR-CE training splits, with 1,000 instructions for reinforcement learning, a 4B GlanceNav reaches success rates of 71.9\% and 73.6\% on R2R-CE and RxR-CE val-unseen. On R2R-CE, it outperforms the strongest previous single-RGB method trained without external data by 9.1 points, while running at 104\,ms per step.

%% file: fig/teaser.tex
\includegraphics[width=\linewidth]{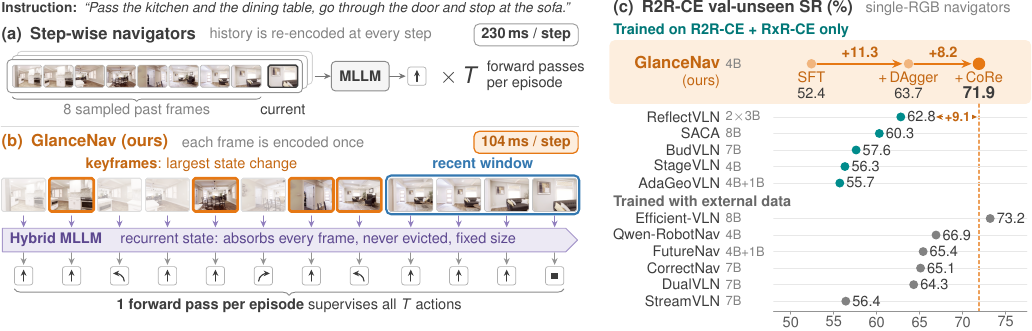}
\caption{\textbf{\method at a glance.}
\textbf{(a)}~Step-wise navigators repeatedly encode sampled historical frames at each decision.
\textbf{(b)}~\method encodes each frame once, maintaining a recurrent state alongside a fixed-size window of recent observations and selected keyframes. This design enables episode-level training with the same memory mechanism used at inference. Step latency is measured on an RTX PRO 6000, using StageVLN~\cite{dao2026stagevln} with a similarly sized backbone as the baseline.
\textbf{(c)}~Success rates (SR) of single-RGB navigators on R2R-CE val-unseen (see \cref{tab:main}). Arrows indicate improvements across our three training stages. $+$1B denotes an additional geometry encoder.
\label{fig:teaser}}

%% file: sec/1_intro.tex
\section{Introduction}
\label{sec:intro}
Vision-and-language navigation in continuous environments (VLN-CE)~\cite{anderson2018vision,krantz2020beyond,ku2020room} asks an agent to follow a natural-language instruction through an unseen building, using only its egocentric observations.
Multimodal large language models (MLLMs) now power most strong VLN-CE agents~\cite{zhang2024navid,zhang2024uni,cheng2024navila,wei2026streamvln,zeng2026janusvln}.
Many of them treat navigation as video question answering (\cref{fig:teaser}a): at every step, they sample a fixed number of past frames, typically eight, and encode them again together with the current view.

This design has two problems.
The first is the memory: each frame is processed once for every later decision that samples it, the policy cannot see what falls outside the sample, and every decision is trained as a separate sample.
Streaming navigators avoid re-encoding by caching the key-value (KV) states of past observations~\cite{wei2026streamvln,zheng2025efficient}.
A full cache, however, grows linearly with the episode, so these methods bound it with sliding windows and pruned or compressed memories.
Bounding the cache fixes the cost but not the memory: eviction is final, and whatever leaves the cache is invisible to every later decision.
The second problem is how the agent learns from its own mistakes.
Beyond DAgger~\cite{ross2011reduction}, recent navigators are fine-tuned with outcome-based reinforcement learning (RL) such as GRPO~\cite{shao2024deepseekmath,zhang2026activevln}, which gives one scalar to all of the roughly one hundred decisions of a rollout and learns mostly from instructions on which some rollouts succeed and others fail, which become rarer as the policy improves.
We therefore ask: \emph{can a navigator encode each frame only once, at a constant cost per step, and still remember what matters and learn efficiently from its own mistakes?}

Hybrid MLLMs~\cite{yang2025gated, hatamizadeh2026gated, qwen35blog} remove the need to choose between cost and memory.
Most of their layers are Gated DeltaNet (GDN) layers, which fold every token into a fixed-size recurrent state, and only a few are softmax attention.
The whole episode is therefore always summarized in the state. The summary is lossy, so the few attention layers still need exact tokens, but only as a complement.
The memory question thus changes from which part of the history to keep to which few frames should \emph{also} be kept exactly.

\method (\cref{fig:teaser}b) answers this question with the state itself.
Its attention layers see a window of the most recent observations and a small bank of older \emph{keyframes}, selected by a \emph{state-change score}: the relative change that a frame causes in the recurrent state of the GDN layer below. A frame that the state already predicts changes it little, whereas a frame with new content, such as a new room, changes it more and is kept.
The score has no parameters.
Two properties follow.
The memory has a constant size: each frame is encoded once, and a step costs the same throughout an episode.
It is also causal: keyframes are selected from the past only, so the memory of every step can be built inside one forward pass over an episode, and \method is trained on whole episodes with its test-time memory instead of on isolated decisions.

The constant size also addresses the second problem: a snapshot of the agent's entire memory takes 136\,MiB and 1.4\,ms to copy, at any step, so a rollout can be rewound to any earlier decision.
Our Counterfactual Rewind (\core) detects when a rollout starts to move away from the goal, restores the simulator and the agent's memory at the rollout's closest approach to the goal, lets a shortest-path expert redo a few actions, and lets the policy finish.
The difference in outcome between this branch and the original rollout then credits the policy's continuation, and the expert's actions only when the branch did better.
A drifting rollout thus yields a contrast between two outcomes from the same state, instead of one scalar for the whole episode.

We train \method in three stages, using only the R2R-CE and RxR-CE training splits: episode-level supervised fine-tuning (SFT), DAgger, and \core on 1{,}000 instructions.
With a 4B backbone, \method reaches 71.9\% success rate (SR) on R2R-CE val-unseen and 73.6\% on RxR-CE (\cref{fig:teaser}c, \cref{tab:main}).
On R2R-CE, this is 9.1 points above the best previous single-RGB method trained without external data, and among all single-RGB methods that we compare with, only an 8B model trained with additional data scores higher.
\method runs at 104\,ms per step, less than half the step time of a step-wise navigator with a backbone of the same size (\cref{tab:speed}).

Our contributions are:
\begin{itemize}[leftmargin=1.5em,itemsep=2pt,parsep=0pt,topsep=3pt]
\item We propose a constant-size streaming memory for hybrid MLLM navigators, which keeps keyframes selected by a parameter-free state-change score, and an episode-level training scheme that matches deployment and encodes $8.6\times$ fewer frames than step-wise training.
\item We propose Counterfactual Rewind, an RL method that branches from a rollout's closest approach to the goal using an $\mathcal{O}(1)$ memory snapshot, and credits the expert's corrections and the policy's continuation by their counterfactual outcome. One pass over 1{,}000 instructions adds 8.2 SR to the DAgger-trained policy, against 3.1 for GRPO on the same instructions (\cref{sec:rl}).
\item With a 4B backbone and no external training data, \method reaches 71.9\% and 73.6\% SR on R2R-CE and RxR-CE val-unseen at a constant 104\,ms per step.
\end{itemize}

%% file: sec/2_related.tex
\section{Related Work}
\label{sec:related}

\para{Vision-language models for navigation.}
Early VLN-CE agents rely on panoramas, depth and waypoint predictors~\cite{krantz2020beyond,krantz2021waypoint,hong2022bridging,wang2023gridmm}.
Video MLLMs made navigation from a single RGB camera competitive~\cite{zhang2024navid,zhang2024uni,cheng2024navila}.
Later work scales the training data~\cite{zhang2026embodied,zhang2026qwen}, or adds self-correction data~\cite{yu2026correctnav}, dual-system planners~\cite{wei2026ground,xue2026omninav}, geometry encoders~\cite{zeng2026janusvln,zhang2026futurenav,pham2026adageovln}, auxiliary objectives~\cite{zhu2026textsc,dao2026stagevln,wang2026think}, agentic decomposition~\cite{xin2026agentvln} or stereo input~\cite{wang2026limits}.
Most of these methods build each decision from a resampled set of frames, and their best results use external data such as ScaleVLN~\cite{wang2023scaling}, VQA or web video.
\method keeps one persistent stream and uses no external data.

\para{Streaming and selective memory.}
StreamVLN~\cite{wei2026streamvln} keeps a sliding window of dialogue turns and a pruned memory of older ones. JanusVLN~\cite{zeng2026janusvln} keeps the KV caches of the initial and the most recent frames from a semantic and a geometry encoder~\cite{wang2025vggt}.
Efficient-VLN~\cite{zheng2025efficient} reuses the KV cache as well and packs several decisions of a trajectory into one training sequence.
In LLMs and video models, KV-eviction methods keep tokens by recency, attention mass or query relevance~\cite{xiao2024efficient,zhang2023h2o,li2024snapkv}.
Linear-attention and state-space layers replace the cache with a fixed-size state~\cite{schlag2021linear,gu2023mamba,dao2024transformers,yang2024parallelizing,yang2025gated}, and hybrid models interleave them with a few attention layers~\cite{team2025kimi,qwen35blog}.
Two recent works read the delta rule to decide which exact tokens to keep: HOLA~\cite{cui2026hippocampus} keeps tokens with a large committed write, and the concurrent DeltaS~\cite{kwon2026deltas} evicts video chunks by their state drift without training.
Our state-change score belongs to this family. Unlike these works, which target language and video understanding, we train a navigation policy end to end with the resulting memory and evaluate it in closed-loop control.

\para{Learning from policy errors.}
DAgger~\cite{ross2011reduction} labels the states visited by the learner and is widely used by MLLM navigators~\cite{wei2026streamvln,zeng2026janusvln,zheng2025efficient}.
ActiveVLN~\cite{zhang2026activevln} trains an MLLM navigator with multi-turn GRPO~\cite{shao2024deepseekmath}, which we adopt as our RL baseline.
Recent methods make its sparse outcome reward more informative: SACA~\cite{Li2026LetsRS} resamples from a divergence point found by an auditor, SeekVLN~\cite{Wang2026SeekBY} contrasts counterfactual branches at evidence-seeking steps, and BudVLN~\cite{He2026NippingTD} rectifies failures from the farthest progress along the reference path.
Branching from intermediate states is also used in LLM reasoning~\cite{Kazemnejad2024VinePPORC}, exploration~\cite{Ecoffet2020FirstRT} and test-time backtracking for VLN~\cite{Ke2019TacticalRS}.
\core branches at the closest approach to the goal, restores the full streaming memory there in $\mathcal{O}(1)$, and imitates the expert only when its branch outperforms the original rollout, a Q-filter~\cite{Nair2017OvercomingEI} estimated by a rollout instead of a learned critic.

%% file: sec/3_method.tex
\section{Method}
\label{sec:method}

\subsection{Preliminaries}
\label{sec:prelim}

\para{Task Formulation.}
Given a natural-language instruction $I$, the agent navigates using egocentric RGB observations $o_t$ received at each time step $t$. It selects an action $a_t$ from $\mathcal{A}=\{\texttt{FORWARD}, \texttt{LEFT}, \texttt{RIGHT}, \texttt{STOP}\}$ according to a policy $\pi_\theta(a_t\mid I, o_{1:t},a_{1:t-1})$. Navigation succeeds when the agent issues \texttt{STOP} within a predefined radius of the goal. We consider an RGB-only setting without access to depth, odometry, or panoramic observations.

\begin{figure}[t]
\centering
\input{fig/method}
\end{figure}

\para{Backbone.}
We build on a hybrid MLLM~\cite{qwen35blog}. A frozen vision encoder maps each frame to $N_v$ visual tokens. The language model stacks $N$ blocks, each made of several GDN layers~\cite{yang2025gated} followed by one softmax-attention layer, so every attention layer sits directly above a GDN layer.
Each GDN head keeps a matrix state $S\in\R^{d_k\times d_v}$ and updates it at every token $i$ with the gated delta rule
\begin{equation}
S_i=\alpha_i S_{i-1}+\beta_i\,k_i e_i^{\top},\qquad e_i=v_i-\alpha_i S_{i-1}^{\top}k_i,
\label{eq:gdn}
\end{equation}
where $\alpha_i\in(0,1)$ is a data-dependent decay, $\beta_i\in(0,1)$ a write strength, $k_i$ a normalized key, and $e_i$ the part of the value $v_i$ that the decayed state does not predict.

\para{Token stream.}
An episode is one token stream $x=[\pre;\grp_1;\grp_2;\dots]$.
The prefix $\pre$ holds the system prompt and the instruction.
The step group $\grp_t$ holds the visual tokens of $o_t$ and the response $r_t$, \ie the action as text.
At test time, $r_t$ is decoded greedily and appended to the stream, so the agent sees its own past actions.

\subsection{Streaming memory}
\label{sec:memory}

At step $t$, only the tokens of $\grp_t$ pass through the model.
What they can see is defined by a memory with three parts (\cref{fig:method}a).

\para{Recurrent state.}
The GDN states are carried through the whole episode and are never evicted. They summarize every observation in a fixed-size state, whatever the episode length.

\para{Recent window.}
Each attention layer keeps the exact KV rows of $\pre$ and of the $W$ most recent groups, with their visual and action tokens.

\para{Keyframe bank.}
When a group leaves the window, its action tokens are dropped and its $N_v$ visual rows become a candidate for the bank $\bnk^{(\ell)}$ of each attention layer $\ell$, which holds at most $F$ frames.
We rank frames by how much they changed the recurrent state of the GDN layer directly below $\ell$ (\cref{fig:method}b).
Let $[b_s,e_s]$ be the token span of observation $s$ and $S_i$ the state of that GDN layer (all heads) after token $i$. The \emph{state-change score} of frame $s$ is
\begin{equation}
\delta^{(\ell)}_s=\frac{\norm{S_{e_s}-S_{b_s-1}}_F}{\max\big(\norm{S_{b_s-1}}_F,\ \varepsilon_\ell\big)},
\label{eq:delta}
\end{equation}
where $\varepsilon_\ell$ is a small constant that only matters for the first frames, when the state is still small.
The score is computed once, when the frame is encoded.
The bank keeps the $F$ frames of $\bnk^{(\ell)}\cup\{s\}$ with the highest scores, and the other frame is deleted.
Kept rows retain their original positions, so nothing is re-encoded.
A query of step $t$ in attention layer $\ell$ attends to
\begin{equation}
\vis^{(\ell)}_t=\pre\,\cup\,\bnk^{(\ell)}_t\,\cup\,\grp_{t-W+1}\cup\dots\cup\grp_{t},
\label{eq:visible}
\end{equation}
causally within $\grp_t$. Each attention layer has its own bank, because useful frames may differ across depth.

\para{What the score measures.}
Unrolling \cref{eq:gdn} over a frame gives, for each head,
\begin{equation}
S_{e}-S_{b-1}=\underbrace{(A-1)\,S_{b-1}}_{\text{forgetting}}+\underbrace{\textstyle\sum_{i=b}^{e}\bar\alpha_i\,\beta_i\,k_i e_i^{\top}}_{\text{new writes}},
\label{eq:decomp}
\end{equation}
with $A=\prod_{i=b}^{e}\alpha_i$ and $\bar\alpha_i=\prod_{j=i+1}^{e}\alpha_j$.
A frame scores high if the state could not predict it (large residuals $e_i$ are written) or if it makes the state forget, as at a room transition. A frame that the state already explains scores low.
The score measures the net effect of both terms, a heuristic for how much new content a frame carries.

\para{Constant cost.}
Each step processes only the tokens of one group. Attention reads at most $|\pre|+F N_v+W\,g_{\max}$ rows per layer, where $g_{\max}$ is the size of the largest group, and the GDN state has a fixed size.
The cost per step therefore does not grow with $t$ (\cref{tab:speed}).

\para{Inference.}
\Cref{alg:step} summarizes one decision of \method.

\begin{figure}[t]
  \begin{minipage}[t]{0.50\linewidth}\vspace{0pt}\input{fig/alg_step}\end{minipage}\hfill
\figpanel[0.47]{fig/mask}
\end{figure}

\subsection{Training}
\label{sec:training}

\para{Episode-level training.}
The scores of a frame depend only on tokens up to that frame, and the bank is a deterministic function of past scores.
The visible sets $\vis^{(\ell)}_t$ of all steps can therefore be computed before attention layer $\ell$ runs.
We serialize a full expert episode with teacher-forced responses $r_t^\star$ and run a single forward pass.
A detached side pass computes the scores and the bank of every step, and each attention layer applies the block-sparse mask $M_{ij}=\mathbf{1}[j\in\vis^{(\ell)}_{t(i)}\wedge j\le i]$, where $t(i)$ is the step of token~$i$ (\cref{fig:mask}).
The loss averages the cross-entropy of the $T$ responses:
\begin{equation}
\mathcal{L}_{\mathrm{SFT}}=-\frac{1}{T}\sum_{t=1}^{T}\frac{1}{|r^\star_t|}\sum_{j=1}^{|r^\star_t|}\log\pi_\theta\big(r^\star_{t,j}\mid x_{<(t,j)};M\big).
\label{eq:sft}
\end{equation}
Gradients flow through every GDN recurrence and through attention over the selected rows. Only the discrete selection is not differentiated.
Unlike step-wise training, which builds one sample per decision and re-encodes its sampled history, one forward pass supervises every decision of an episode and encodes each frame once (\cref{tab:paradigm}).

\para{DAgger.}
To expose the policy to its own mistakes, we follow common practice \cite{wei2026streamvln,zeng2026janusvln} and collect on-policy data with the Stage-1 policy. A shortest-path expert takes over after clear failures.
We keep the executed actions in the input stream, so that the memory is that of the rollout, but label every visited state with the expert's action and train on these episodes together with the expert demonstrations (\cref{sec:ablation}).

\subsection{Counterfactual Rewind}
\label{sec:core}

\para{Base algorithm.}
We follow the multi-turn GRPO recipe of ActiveVLN~\cite{zhang2026activevln}.
For each instruction, the policy samples $G$ \emph{main} rollouts $\tau^{1:G}$. Rollout $\tau^g$ receives a terminal reward $R^g$, which is zero unless the agent stops within the success radius and otherwise decreases linearly with its final distance to the goal. All its decisions share the advantage $\hat A^{g}=(R^g-\mathrm{mean}\,R)/(\mathrm{std}\,R+\epsilon)$ within the group.
In navigation, this signal is coarse, since one scalar covers every decision of a long episode. It also vanishes when all rollouts of a group fail, and merely ranks their stop distances when all succeed; such groups become common as the policy improves (\cref{sec:rl}).

\para{Closest approach.}
Let $d_k$ be the geodesic distance to the goal before decision $k$, which the simulator provides, as for the reward.
A main rollout is \emph{triggered} at the first decision at which it is a margin $\eta$ farther from the goal than its best so far (\cref{fig:method}c):
\begin{equation}
\begin{aligned}
t^\dagger&=\min\{t:\ d_t\ge \min_{k<t}d_k+\eta\},\\
k^\star&=\argmin_{k<t^\dagger}d_k.
\end{aligned}
\label{eq:trigger}
\end{equation}
Decision $k^\star$ is the rollout's closest approach to the goal.

\para{Counterfactual branch.}
While a main rollout runs, we copy the agent's memory whenever $d_k$ reaches a new minimum, until the rollout is triggered. Because the memory has a constant size, a copy is cheap and costs the same at every step.
If the rollout is triggered, we restore the simulator pose and the memory at $k^\star$ after it ends.
A shortest-path expert then takes at most $H$ actions, which are written into the agent's own memory as its responses, and the policy continues with fresh samples for as many decisions as the main rollout had left at $k^\star$.
Let $R_b$ be the return of branch $b$ and $R_{m}$ that of its main rollout.

\para{Counterfactual credit.}
The branch is scored against its main rollout, $\Delta_b=(R_b-R_m)/s$, where $s=\max(\sigma_R,s_{\min})$, $\sigma_R$ is the standard deviation of the main returns in the batch, and the floor $s_{\min}$ prevents division by a vanishing spread.
Each response token $(t,j)$ then receives a weight $w_{t,j}$. Tokens of a main rollout $\tau^g$ keep its advantage $\hat A^g$. In a branch, the policy's continuation receives $\Delta_b$, and the expert's actions receive $[\Delta_b]_+=\max(0,\Delta_b)$, so a correction is imitated only if its branch obtained a higher return than the main rollout, a counterfactual Q-filter~\cite{Nair2017OvercomingEI}. The prefix that the branch shares with its main rollout receives $w_{t,j}=0$.
$\Delta_b$ contrasts two outcomes from the same state: it credits the correction and the continuation as a whole, not each action.
The objective is
\begin{equation}
\mathcal{L}_{\mathrm{CoRe}}=-\frac{1}{Z}\sum_{(t,j)}w_{t,j}\,\log\pi_\theta\big(y_{t,j}\mid x_{<(t,j)}\big),
\label{eq:core}
\end{equation}
where $y_{t,j}$ is response token $j$ at decision $t$, conditioned on the full stream $x_{<(t,j)}$ before it, and $Z$ is the number of response tokens of the main rollouts, as in the baseline.
We optimize it with the clipped surrogate of the GRPO baseline and one on-policy update per batch, so that the importance ratio is one and the gradient equals that of \cref{eq:core}.
\Cref{eq:core} thus combines a policy gradient on sampled tokens with outcome-filtered imitation of the expert's. Each triggered rollout has one branch. If no rollout is triggered, \cref{eq:core} is the GRPO objective, so \core changes training only where a rollout drifts.
\Cref{alg:core} summarizes one \core update.

\begin{figure}[t]
\centering
\begin{minipage}{0.80\linewidth}
\input{fig/alg_core}
\end{minipage}
\end{figure}

\para{Validity of the credit.}
The return of the main rollout does not depend on the actions sampled in the branch, so $R_m$ is a valid baseline for the policy's continuation.
The expert's tokens are not samples of the policy, so their term is imitation, weighted by the outcome of the branch. The weight $[\Delta_b]_+$ is shared by the whole correction: a helpful correction is not credited if the continuation fails by chance, and each of its actions is credited when the branch as a whole does better.

\para{Why the closest approach?}
At the trigger $t^\dagger$, the agent is already lost, and most expert actions there are turns toward a goal that the agent cannot see. At $k^\star$, the expert instead redoes the moves with which the agent left its best position (\cref{sec:rl}).
Distances and expert actions are used only during training; the policy needs neither at test time.

%% file: fig/method.tex
\includegraphics[width=\linewidth]{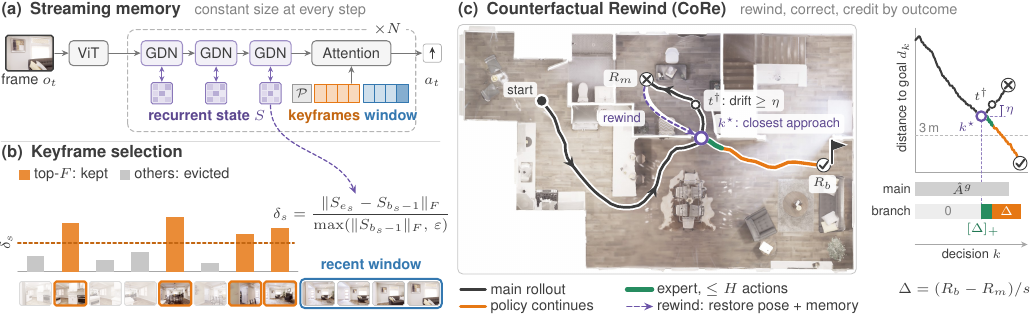}
\caption{\textbf{Overview of \method.}
\textbf{(a)}~Each frame is encoded once. GDN layers update recurrent states, while attention retains the instruction prefix $\pre$, recent steps, and selected keyframes within a fixed memory budget.
\textbf{(b)}~Frames are scored once by their relative recurrent-state change $\delta_s$ (\cref{eq:delta}). On leaving the recent window, they compete for a layer-specific top-$F$ bank.
\textbf{(c)}~\core (training only). After the main rollout ends, \core restores the pose and full memory at $k^\star$, the closest checkpoint before the first drift trigger $t^\dagger$. An expert supplies at most $H$ corrective actions before the policy resumes. The branch--main return difference, $R_b-R_m$, determines positive-only expert credit and signed policy credit.}
\label{fig:method}

%% file: fig/alg_step.tex
\begin{algorithm}[H]
\caption{One \method decision at step $t$}
\label{alg:step}
\small
\begin{algorithmic}[1]
\Require GDN states, per-layer KV rows of $\pre$, $\bnk^{(\ell)}$ and $\grp_{t-W:t-1}$; image $o_t$
\If{$t>W$} \Comment{step $s=t{-}W$ leaves the window}
  \For{each attention layer $\ell$}
    \State $\bnk^{(\ell)}\gets\mathrm{TopF}_{\delta^{(\ell)}}\big(\bnk^{(\ell)}\cup\{\text{visual rows of }\grp_s\}\big)$
    \State delete the action rows of $\grp_s$ and the rejected frame
  \EndFor
\EndIf
\State encode $o_t$ with the vision encoder \Comment{once}
\State append the observation tokens through all layers; GDN states update, attention reads $\vis^{(\ell)}_t$ (\cref{eq:visible})
\State store $\delta^{(\ell)}_t$ for all attention layers (\cref{eq:delta})
\State decode $r_t$ greedily, append it, \Return $a_t=\mathrm{parse}(r_t)$
\end{algorithmic}
\end{algorithm}

%% file: fig/alg_core.tex
\begin{algorithm}[H]
\caption{One \core update}
\label{alg:core}
\small
\begin{algorithmic}[1]
\Require policy $\pi_\theta$, instructions $\{I_n\}$, group size $G$
\For{each $I_n$ and $g=1,\dots,G$}
  \State roll out $\tau^g\sim\pi_\theta$; until the trigger, copy memory and pose at each new minimum of $d_k$ \Comment{$\mathcal{O}(1)$}
  \State compute $R^g$; if triggered (\cref{eq:trigger}), the last copy is the one at $k^\star$
\EndFor
\State re-roll all-fail groups, then compute $\hat A^{g}$
\For{each triggered main rollout $m$}
  \State restore pose and memory at $k^\star$
  \State expert acts for at most $H$ actions, written into memory
  \State policy continues; $\Delta_b\gets(R_b-R_m)/s$
\EndFor
\State one update on \cref{eq:core}: mains $\hat A$, expert $[\Delta_b]_+$, continuation $\Delta_b$
\end{algorithmic}
\end{algorithm}

%% file: sec/4_experiments.tex
\section{Experiments}
\label{sec:exp}

\begin{table}[t]
\centering
\input{tab/main}
\end{table}

\subsection{Experimental Setup}
\label{sec:setup}

\para{Benchmarks and metrics.}
We evaluate on the val-unseen splits of R2R-CE~\cite{krantz2020beyond} (1{,}839 episodes) and RxR-CE~\cite{ku2020room} (3{,}669 English episodes) in Habitat~\cite{savva2019habitatplatformembodiedai} with Matterport3D scenes~\cite{chang2017matterport3dlearningrgbddata}.
The agent observes $480{\times}640$ RGB images, moves forward 0.25\,m or turns $15^\circ$ per action, and succeeds if it stops within 3\,m of the goal.
Following standard practice, we report navigation error (NE), oracle success (OS), success rate (SR) and success weighted by path length (SPL)~\cite{anderson2018evaluationembodiednavigationagents}, and on RxR-CE also normalized dynamic time warping (nDTW)~\cite{ilharco2019generalevaluationinstructionconditioned}.

\para{Model.}
We instantiate \method with Qwen3.5-4B~\cite{qwen35blog}, whose language model has $N{=}8$ blocks of three GDN layers and one attention layer. The frozen vision encoder maps each frame to $N_v{=}300$ tokens.
The memory keeps $W{=}4$ recent steps and $F{=}4$ keyframes per attention layer, so attention reads at most 4{,}784 KV rows per layer, and the recurrent state occupies 48\,MiB.

\para{Training.}
All three stages use only the R2R-CE and English RxR-CE training splits (30{,}815 episodes with 3.13M actions).
Stage~1 trains for one epoch, which takes 13.5\,h on four H100 GPUs.
Stage~2 collects DAgger episodes with the Stage-1 policy on the same instructions and trains for one more epoch with the same recipe.
Stage~3 runs \core on 1{,}000 training instructions, 500 from R2R-CE and 500 from RxR-CE.
Each update uses $G{=}4$ rollouts per instruction, and one pass over the 1{,}000 instructions takes 125 updates, or 12.5\,h on two RTX PRO 6000 GPUs.
\core uses a trigger margin $\eta{=}1$\,m and at most $H{=}16$ expert actions.
Full hyperparameters and data-collection settings will be released with the code.

\para{Compared methods.}
We report published results, and group the methods by their training data (\cref{tab:main}).
Most methods add data from other sources, such as ScaleVLN~\cite{wang2023scaling}, VQA or web video.
The others use only the R2R-CE and RxR-CE training splits, either as demonstrations or, like \method, also for DAgger or RL.

\subsection{Comparison with the State of the Art}
\label{sec:main}

\para{Navigation performance.}
\method reaches 71.9 SR and 63.7 SPL on R2R-CE val-unseen (\cref{tab:main}).
Among single-RGB methods that use only the R2R-CE and RxR-CE training splits, the best previous SR that we are aware of is 62.8 (ReflectVLN~\cite{wang2026reflectvln}, which runs two 3B models and trains on recovery data), and the best SR from demonstrations alone is 56.3 (StageVLN~\cite{dao2026stagevln}).
\method improves on them by 9.1 and 15.6 points with a 4B backbone.
Among all single-RGB methods of \cref{tab:main}, only Efficient-VLN~\cite{zheng2025efficient}, an 8B model trained with ScaleVLN data, has a higher SR (73.2), and \method has the highest OS (78.6).
On RxR-CE, \method{} reaches 73.6 SR, 7.6 points above ReflectVLN, and again only Efficient-VLN is higher (75.6).

\begin{table}[t]
\tabpanel[0.50]{tab/speed}\hfill
\tabpanel[0.47]{tab/paradigm}
\end{table}

\para{Inference cost.}
\Cref{tab:speed} follows the protocol of StageVLN~\cite{dao2026stagevln}.
On an RTX PRO 6000, \method takes 103.9\,ms and 8.68\,GiB per step, about half the time and memory of StreamVLN and less than half the time of StageVLN, which re-encodes eight frames with a backbone of the same size.
Because the memory has a constant size, the step time stays flat along an episode. With a full KV cache instead, the same backbone slows from 93.6 to 152.8\,ms per step over 500 steps, while an eight-step window stays at 84\,ms (\cref{tab:fullcontext}).
For the same reason, a snapshot of the full memory, as used by \core, takes 136\,MiB and 1.4\,ms at any step.

\para{Latency breakdown.}
In an instrumented run on the RTX PRO 6000, a step spends 8.7\,ms in the vision encoder, 35.5\,ms appending the observation, and 55.4\,ms decoding the response.
Window-only models ($W{=}4$ to $9$) run at 84--87\,ms per step and bank models at 103--106\,ms, for every bank size we measured. A randomly filled bank of the same size runs at 89.1\,ms, so about 15\,ms go to our unfused scoring code, which could be fused into the GDN kernel.

\begin{table}[t]
\centering
\input{tab/fullcontext}
\end{table}

\para{Memory snapshot.}
A snapshot holds the recurrent and convolution states of the 24 GDN layers (49.5\,MiB in FP32), the KV rows of prefix, bank and window of the 8 attention layers, the bank scores and the step counter: 135.7\,MiB in total, cloned on the GPU in 1.4\,ms.
These numbers are measured on training episodes, whose instruction prefixes are far shorter than the maximum prefix length.
A full-history KV cache of the same layers would need at least 914\,MiB at $t{=}100$ and 3{,}502\,MiB at $t{=}400$, and a re-encoding policy would have to re-encode its history after each rewind.

\para{Training cost.}
\Cref{tab:paradigm} compares episode-level training with the step-wise training used by most MLLM navigators, on the same backbone and data.
The step-wise model encodes, for every decision, the current frame and up to eight sampled history frames, 27.0M frames per epoch. The episode-level model encodes each of the 3.13M frames once, $8.6\times$ fewer, and supervises about 100 decisions per forward pass, so one epoch takes 13.5\,h instead of 42.6\,h.
It also reaches a higher SR (52.4 \vs 46.5), although the step-wise model comes from a separate code base, so this compares two systems and does not isolate the training scheme.

\para{Agreement of training and inference.}
We replay real episodes of 64 and 128 steps with the same actions through the training pass and through the streaming path of the Stage-1 model.
The two paths agree on the most likely response token in 99.5 to 100\% of the cases and on 99.4 to 99.7\% of the bank rows, and their logits differ by a root mean square of 0.03.
The remaining differences are consistent with the different numerical kernels that the two paths use for the GDN recurrence.

\subsection{Ablation Study}
\label{sec:ablation}

\begin{table}[t]
\tabpanel[0.53]{tab/memory}\hfill
\figpanel[0.44]{fig/budget}
\end{table}

\begin{table}[t]
\tabpanel[0.46]{tab/randombank}\hfill
\tabpanel[0.50]{tab/budget}
\end{table}

\para{Streaming memory.}
\Cref{tab:memory} trains a Stage-1 model for each memory.
Adding a bank of four keyframes to the four-step window raises SR from 48.7 to 52.4.
The gain depends on what the bank holds. Averaging the token scores of a frame gives 49.3, close to the window alone, and under an earlier training schedule, a bank of randomly chosen tokens did not help at all, whereas a scored bank of the same size raised SR from 46.3 to 49.3 (\cref{tab:randombank}).
Among the better scores, the differences are small. Keeping tokens by their committed write, as in HOLA~\cite{cui2026hippocampus}, or by a learned ranker gives 51.7, and our frame-level score is not significantly better ($p{=}0.59$, exact McNemar test).
We use $\delta_s$ because it keeps whole frames, has no parameters, and has the clear reading of \cref{eq:decomp}.

\para{Scoring rule.}
\Cref{tab:memory} compares ten parameter-free scores read from the GDN layer below each attention layer, and two learned rankers trained to predict a token's future attention mass.
The frame-level state-change score of \cref{eq:delta} gives the highest SR. However, the SR of the ten GDN-derived scores has a standard deviation of only 1.1 points (49.0 to 52.4). Neither learned ranker improves on the parameter-free scores.
Averaging the committed write over a frame is the worst frame-level rule, so the result is not explained by frame granularity alone.

\para{Random bank.}
Under an earlier schedule (the epoch-1 checkpoint of a two-epoch run), we compared random and scored banks directly (\cref{tab:randombank}).
A bank of random old tokens does not help ($-0.2$ against the window alone, $p{=}0.90$), whereas a bank of the same size filled by the committed write raises SR from 46.3 to 49.3 ($p{=}0.023$) and beats the random bank by 3.2 points ($p{=}0.018$).
Moving budget from the window to the bank helps further, and a ninth recent step, which costs about as much as a one-frame bank, lowers SR.
The policy also relies on its bank: serving the committed-write model of \cref{tab:memory} without its bank costs 1.7 points ($p{=}0.12$).

\para{Memory budget.}%
\Cref{tab:budget} trains a model for each budget. With eight visible frames, a window of four steps and four keyframes outperforms a window of eight steps by 3.4 SR on R2R-CE and 1.3 on RxR-CE, and a window alone needs twelve frames to come close (51.2 and 56.7).
On R2R-CE, a larger bank or window does not help further (51.3 and 51.0). On RxR-CE, the banks with $W{=}4$ lie within 1.7 points of each other, and two keyframes score highest.
We keep $W{=}4$ and $F{=}4$, the best budget on R2R-CE.
\Cref{fig:budget} plots the eight models of \cref{tab:budget} against the number of frames visible to an attention layer.
Starting from the four-step window, the next frames help more as keyframes than as recent steps on both benchmarks: with six or eight visible frames, every model with a bank outperforms the window-only models, and a window alone needs twelve frames to be on par.
The split matters as well. A window of two steps with four keyframes is 3.2 points below $W{=}4$ on R2R-CE (0.4 on RxR-CE), and a window of eight steps does not improve on $W{=}4$ either.

\para{DAgger labels and data.}
We follow the recovery protocol of JanusVLN~\cite{zeng2026janusvln}: the Stage-1 policy drives greedily on the training instructions, and a shortest-path expert takes over after clear failures and returns control to the policy at the next waypoint.
Because the memory of \method contains its own responses, a takeover restores the memory snapshot taken after the observation and appends the expert's response instead of the learner's. Inputs, recurrent state and bank are therefore those of the executed trajectory.
We keep episodes that end at the goal and contain a detour, which gives 14{,}195 episodes with 1.61M steps.
In 31\% of the collected steps, the executed action is a mistake of the learner made before a takeover, so supervising the executed actions imitates these mistakes.
We therefore keep the input stream bit-exact, so that recurrent state and bank match the rollout, and replace only the targets by the expert's action at every visited state (\emph{oracle relabeling}).
During collection, the expert also overrides the learner's \texttt{STOP} proposals until the agent is very close to the goal, so near the goal both labels say ``keep moving''. We truncate each episode at the learner's first \texttt{STOP} proposal near the goal and label that step \texttt{STOP} (\emph{STOP correction}).
We train for one epoch from the Stage-1 weights on the DAgger episodes together with the expert demonstrations of the 16{,}620 training instructions for which no DAgger episode was kept, so that every instruction appears once (30{,}815 episodes in total).

\Cref{tab:dagger} shows that both the labels and the demonstrations matter. Supervising executed actions does not improve SR, although the agent reaches the goal area far more often (OS 58.0$\to$70.3), because it also learns the learner's mistakes and the expert's refusals to stop. Oracle relabeling adds 6.8 points with the same inputs.
DAgger episodes alone lead to long paths (15.1\,m). Adding demonstrations restores path efficiency and gives the best model.

\begin{table}[t]
\centering
\input{tab/dagger}
\end{table}

\subsection{Analysis of Counterfactual Rewind}
\label{sec:rl}

\para{Setup.}%
Every RL run starts from the Stage-2 model and makes one pass over 1{,}000 instructions. GRPO and \core see the same instructions in the same order, with the same seed: the first 500 of the R2R-CE list and of the RxR-CE list released with ActiveVLN~\cite{zhang2026activevln}, without any selection. This is 3\% of our training instructions and an eighth of those that ActiveVLN uses. GRPO follows the recipe of ActiveVLN, and \core adds its branches to it. We report one training run per setting.

\para{Training stages.}
\Cref{tab:rl} shows that every stage helps.
DAgger raises SR from 52.4 to 63.7. This requires labeling every visited state with the expert's action; supervising the executed actions instead leaves SR unchanged (\cref{tab:dagger}).
\core then adds 8.2 points in 12.5\,h on two GPUs. RxR-CE improves with every stage as well, from 56.6 to 67.2 and 73.6 SR.

\begin{table}[t]
\tabpanel[0.47]{tab/rl}\hfill
\figpanel[0.50]{fig/rl}
\end{table}

\begin{table}[t]
\tabpanel[0.46]{tab/core}\hfill
\figpanel[0.51]{fig/core_analysis}
\end{table}

\para{\core \vs GRPO.}%
From the same instructions, \core gains more than GRPO in both settings that we ran: 71.9 against 66.8 SR with the mixed instructions (\cref{tab:rl}), and 68.7 against 64.3 with R2R-CE instructions only (\cref{tab:core}).
On the full RxR-CE val-unseen split, GRPO and \core reach 72.4 and 73.6 SR, 59.6 and 59.6 SPL, and 70.3 and 70.7 nDTW. Both improve the Stage-2 model (67.2 SR, 54.8 SPL, 65.7 nDTW) by more than five points of SR, and \core is 1.2 points ahead of GRPO. This margin is smaller than on R2R-CE. A likely reason is that the trigger of \core measures the distance to the goal and its expert follows the shortest path, whereas RxR-CE routes are not shortest paths: a branch can start on a correct detour and teach a shortcut.

\para{Why \core learns more from the same instructions.}%
GRPO learns from the contrast between the rollouts of an instruction, and as the policy improves, fewer groups contain both a success and a failure: 64\% in the first quarter of training, 36\% in the last (\cref{fig:rl}a). All-fail groups carry no signal. In all-success groups, half of all groups, returns differ only through a stop distance that varies by 0.23\,m, yet group normalization gives them half of the total advantage.
A branch contrasts two outcomes from the same state. \core keeps about 60\% of the groups informative, 75\% of its branches succeed, and they rescue 53\% of the failed main rollouts.

\para{Where the gain comes from.}%
In \cref{fig:rl}b, the Stage-2 model stops without ever coming within 3\,m of the goal in 27.0\% of the episodes. \core lowers this share to 17.5\% and GRPO to 20.3\%, while 6 to 8\% of the episodes still reach the goal area and leave it. \core thus mainly helps the agent reach the goal area, and does not change how reliably it stops.

\para{Design of \core.}%
\Cref{tab:core} varies one design choice at a time, with 1{,}000 R2R-CE instructions. Its entries are single runs, and runs of one configuration differ by about two points.
GRPO with a fifth rollout per instruction matches the training time of \core and is 2.2 points worse, so the gain does not come from the additional rollouts.
Where a branch starts matters: at the trigger, 80\% of the expert's actions are turns (57\% at $k^\star$), the policy learns to spin, its paths grow from 11.2 to 13.0\,m, and SR is 2.0 points lower.
Who acts matters less: if the policy resamples from $k^\star$ itself, SR is 1.4 points lower.
The credit on the expert's corrections carries most of the gain: without it, SR falls by 3.8 points, close to GRPO, and imitating every correction regardless of its outcome costs 1.7 points.
Among the trigger margins, $\eta{=}1$\,m is best. A shorter expert horizon changes little (69.1 for $H{=}4$), whereas $H{=}32$ costs 2.2 points.

\para{Design study.}
\Cref{fig:core_analysis}a counts the instruction groups that carry a training signal in the runs of \cref{tab:core}.
As GRPO improves, more groups become all-success or all-failure: over the whole run, about half of its groups have no outcome contrast.
In the first 30 updates of \core, 30\% of the main rollouts trigger a branch and 72.8\% of the branches succeed, rescuing 170 of 338 failed rollouts.
Self-rewind keeps almost as many groups informative as \core, but its branches succeed less often (44.5\%) and rescue 101 of 343 failed rollouts: the expert adds successful corrections, not more contrast.
In \cref{fig:core_analysis}b, the share of episodes that reached the goal area and then left stays at 7--8\% for all models, so \core does not change how reliably the agent stops once it is near the goal.

\para{Where to branch.}
Imitating expert turn-arounds can make the policy spin in place.
When a branch starts at the trigger $t^\dagger$, 80\% of the expert's actions are turns. Rewinding to $k^\star$ lowers this share to 57\% (62\% with the mixed instructions).
In a probe on 16 failed rollouts of the Stage-1 model, branching at $k^\star$ with the expert rescued 12 of them, and branching at $t^\dagger$ only 6 ($p{=}0.07$).
With the mixed instructions and the same seed, rewinding to the farthest progress along the reference path, as in BudVLN~\cite{He2026NippingTD}, gives 69.7 SR, against 71.9 for the closest approach to the goal.

\subsection{Real-World Experiments}
\label{sec:real}

\begin{figure}[t]
\centering
\input{fig/realworld}
\end{figure}

We deploy \method on a Unitree G1 humanoid robot equipped with a chest-mounted ZED X Mini camera, using only the left RGB stream. A workstation with an RTX 5090 runs the navigation policy and receives observations over Wi-Fi. All real-world experiments are conducted without fine-tuning on real-world data.

The instructions fall into two categories: (i) \textbf{T1 (Moderate-length Navigation)}, which requires the robot to distinguish visually similar objects and use their spatial or sequential relationships to identify the correct stopping location; and (ii) \textbf{T2 (Long-range Navigation)}, which requires landmark recognition and navigation decisions at multiple points along the route.\looseness=-1

\Cref{fig:realworld} shows one successful episode of each category, in which the robot recognizes the landmarks and stops at the described location. More real-world results are in \hyperref[sec:supp_real]{App.~\ref*{sec:supp_real}}.

%% file: tab/main.tex
\caption{\textbf{Comparison with state-of-the-art methods on R2R-CE and RxR-CE val-unseen.} Pano., Odo., Depth and S.RGB: panoramic view, odometry, depth and single RGB input. External data: \cmark\ if a method trains on data from sources other than the R2R-CE and RxR-CE training splits, as reported by each method, and \xmark\ if it does not. Methods with other inputs are shown in gray for reference. Among single-RGB methods, the best and second-best results are in \textbf{bold} and \underline{underlined}. $^\dagger$Additional geometry encoder. $^\ddagger$Multi-view RGB.}
\label{tab:main}
\setlength{\tabcolsep}{5.2pt}
\renewcommand{\arraystretch}{0.93}
\footnotesize
\begin{adjustbox}{max width=\linewidth}\begin{tabular}{@{}+l ^c ^c^c^c^c ^c^c^c^c ^c^c^c^c ^c@{}}
\toprule
 & & \multicolumn{4}{c}{Observation} & \multicolumn{4}{c}{R2R-CE val-unseen} & \multicolumn{4}{c}{RxR-CE val-unseen} & External\\
\cmidrule(lr){3-6}\cmidrule(lr){7-10}\cmidrule(lr){11-14}
Method & Backbone & Pano. & Odo. & Depth & S.RGB & NE$\downarrow$ & OS$\uparrow$ & SR$\uparrow$ & SPL$\uparrow$ & NE$\downarrow$ & SR$\uparrow$ & SPL$\uparrow$ & nDTW$\uparrow$ & data\\
\midrule
\multicolumn{15}{@{}l}{\textit{Panoramic, depth or multi-view input (for reference)}}\\
\rf GridMM~\cite{wang2023gridmm}          & -- & \cmark & \cmark & \cmark &        & 5.11 & 61.0 & 49.0 & 41.0 & --   & --   & --   & -- & \cmark\\
\rf Reborn~\cite{an20221stplacesolutionsrxrhabitat} & -- & \cmark & \cmark & \cmark &        & 5.40 & 57.0 & 50.0 & 46.0 & 5.98 & 48.6 & 42.1 & 63.4 & \cmark\\
\rf NavMorph~\cite{yao2025navmorphselfevolvingworldmodel}       & -- &        &        & \cmark & \cmark & 5.75 & 56.9 & 47.9 & 33.2 & 8.85 & 30.8 & 22.8 & 44.2 & --\\
\rf AgentVLN~\cite{xin2026agentvln}       & 3B &        &        & \cmark & \cmark & 3.88 & 73.5 & 67.2 & 64.7 & 3.92 & 69.5 & 61.3 & 74.6 & \cmark\\
\rf OmniNav\smash{$^\ddagger$}~\cite{xue2026omninav} & 3B & \cmark &        &        &        & 3.74 & 74.6 & 69.5 & 66.1 & 3.77 & 73.6 & 62.0 & -- & \cmark\\
\rf Qwen-RobotNav\smash{$^\ddagger$}~\cite{zhang2026qwen} & 8B & \cmark &  &  &  & 3.53 & 78.5 & 72.1 & 66.6 & 3.58 & 76.5 & 65.7 & 72.5 & \cmark\\
\midrule
\multicolumn{15}{@{}l}{\textit{Single RGB, trained with external data}}\\
NaVid~\cite{zhang2024navid}           & 7B &  &  &  & \cmark & 5.47 & 49.1 & 37.4 & 35.9 & 8.41 & 23.8 & 21.2 & -- & \cmark\\
Uni-NaVid~\cite{zhang2024uni}    & 7B &  &  &  & \cmark & 5.58 & 53.3 & 47.0 & 42.7 & 6.24 & 48.7 & 40.9 & -- & \cmark\\
ActiveVLN~\cite{zhang2026activevln} & 3B &  &  &  & \cmark & 5.31 & 58.3 & 50.1 & 43.7 & 5.84 & 50.7 & 41.2 & 58.1 & \cmark\\
NaVILA~\cite{cheng2024navila}         & 8B &  &  &  & \cmark & 5.22 & 62.5 & 54.0 & 49.0 & 6.77 & 49.3 & 44.0 & 58.8 & \cmark\\
InternVLA-N1 (S2)~\cite{wei2026ground} & 7B &  &  &  & \cmark & 4.89 & 60.6 & 55.4 & 52.1 & 6.41 & 49.5 & 41.8 & 62.6 & \cmark\\
NavFoM~\cite{zhang2026embodied}         & 7B &  &  &  & \cmark & 5.01 & 64.9 & 56.2 & 51.2 & 5.51 & 57.4 & 49.4 & 60.2 & \cmark\\
StreamVLN~\cite{wei2026streamvln}     & 7B &  &  &  & \cmark & 4.90 & 63.6 & 56.4 & 50.2 & 5.65 & 54.4 & 45.4 & 63.7 & \cmark\\
JanusVLN\smash{$^\dagger$}~\cite{zeng2026janusvln} & 7B+1B &  &  &  & \cmark & 4.78 & 65.2 & 60.5 & 56.8 & 6.06 & 56.2 & 47.5 & 62.1 & \cmark\\
NaVIDA~\cite{zhu2026textsc}           & 3B &  &  &  & \cmark & 4.32 & 69.5 & 61.4 & 54.7 & 5.23 & 57.4 & 49.6 & 67.0 & \cmark\\
DualVLN~\cite{wei2026ground}         & 7B &  &  &  & \cmark & 4.05 & 70.7 & 64.3 & 58.5 & 4.58 & 61.4 & 51.8 & 70.0 & \cmark\\
SACA~\cite{Li2026LetsRS}                & 8B &  &  &  & \cmark & 4.19 & 69.3 & 64.7 & 56.9 & 4.75 & 62.1 & 51.7 & 66.0 & \cmark\\
CorrectNav~\cite{yu2026correctnav}    & 7B &  &  &  & \cmark & 4.24 & 67.5 & 65.1 & 62.3 & 4.09 & 69.3 & \underline{63.3} & \textbf{75.2} & \cmark\\
FutureNav\smash{$^\dagger$}~\cite{zhang2026futurenav} & 4B+1B &  &  &  & \cmark & 4.24 & 70.2 & 65.4 & 61.3 & 4.60 & 60.9 & 52.8 & 68.0 & \cmark\\
Qwen-RobotNav~\cite{zhang2026qwen} & 4B &  &  &  & \cmark & 4.22 & 73.6 & 66.9 & 60.5 & 4.15 & 71.3 & 61.5 & 68.6 & \cmark\\
Qwen-RobotNav~\cite{zhang2026qwen} & 8B &  &  &  & \cmark & 4.36 & 72.7 & 65.7 & 59.6 & 4.16 & 73.4 & \textbf{63.5} & 69.9 & \cmark\\
SeekVLN~\cite{Wang2026SeekBY} & 8B &  &  &  & \cmark & 3.7\phantom{0} & 75.2 & 67.5 & 61.4 & 4.9\phantom{0} & 59.7 & 50.3 & 63.6 & \cmark\\
Efficient-VLN~\cite{zheng2025efficient} & 8B &  &  &  & \cmark & \textbf{3.29} & \underline{78.2} & \textbf{73.2} & \textbf{65.0} & \textbf{3.18} & \textbf{75.6} & 60.0 & -- & \cmark\\
\midrule
\multicolumn{15}{@{}l}{\textit{Single RGB, R2R-CE and RxR-CE only: expert demonstrations}}\\
JanusVLN\smash{$^\dagger$}~\cite{zeng2026janusvln} & 7B+1B &  &  &  & \cmark & 5.17 & 58.0 & 52.8 & 49.2 & 6.46 & 51.4 & 44.3 & 59.1 & \xmark\\
FutureNav\smash{$^\dagger$}~\cite{zhang2026futurenav} & 4B+1B &  &  &  & \cmark & 5.13 & 61.8 & 55.1 & 50.1 & 5.62 & 54.5 & 46.0 & 61.8 & \xmark\\
AdaGeoVLN\smash{$^\dagger$}~\cite{pham2026adageovln} & 4B+1B &  &  &  & \cmark & 5.27 & 60.7 & 55.7 & 51.4 & 5.64 & 54.1 & 44.7 & 61.8 & \xmark\\
StageVLN~\cite{dao2026stagevln}       & 4B &  &  &  & \cmark & 5.17 & 64.7 & 56.3 & 51.4 & 5.65 & 54.3 & 42.8 & 58.3 & \xmark\\
\midrule
\multicolumn{15}{@{}l}{\textit{Single RGB, R2R-CE and RxR-CE only: demonstrations and on-policy training (DAgger, recovery data or RL)}}\\
MonoDream~\cite{wang2026monodream} & 2B &  &  &  & \cmark & 5.45 & 61.5 & 55.8 & 49.1 & 6.38 & 49.4 & 40.9 & -- & \xmark\\
BudVLN~\cite{He2026NippingTD}            & 7B &  &  &  & \cmark & 4.74 & 65.6 & 57.6 & 51.1 & 5.79 & 56.1 & 46.6 & 63.2 & \xmark\\
SACA~\cite{Li2026LetsRS}                & 8B &  &  &  & \cmark & 4.57 & 64.9 & 60.3 & 55.1 & 4.90 & 60.3 & 49.8 & 62.1 & \xmark\\
ReflectVLN~\cite{wang2026reflectvln} & 2$\times$3B &  &  &  & \cmark & 4.19 & 67.3 & 62.8 & 58.5 & 3.98 & 66.0 & 57.2 & -- & \xmark\\
\rowcolor{oursbg}\method (ours)       & 4B &  &  &  & \cmark & \underline{3.32} & \textbf{78.6} & \underline{71.9} & \underline{63.7} & \underline{3.55} & \underline{73.6} & 59.6 & \underline{70.7} & \xmark\\
\bottomrule
\end{tabular}\end{adjustbox}

%% file: tab/fullcontext.tex
\begin{sidecap}[0.44]
\caption{\textbf{Full KV cache \vs bounded window.} Median step time (ms) of a full KV cache and an eight-step window over 500 measured steps after 16 warm-up steps, in bins of 100 steps (RTX PRO 6000, same backbone). KV rows: the largest number of rows of an attention layer in each bin.}
\label{tab:fullcontext}
\end{sidecap}
\begin{sidebody}[0.52]
\setlength{\tabcolsep}{4pt}
\footnotesize
\begin{adjustbox}{max width=\linewidth}\begin{tabular}{@{}lccc@{}}
\toprule
Steps & Full cache & Window ($W{=}8$) & Full-cache KV rows\\
\midrule
16--115 & 93.6 & 84.0 & 36{,}962\\
116--215 & 110.4 & 84.1 & 68{,}762\\
216--315 & 120.4 & 84.5 & 100{,}562\\
316--415 & 138.1 & 84.1 & 132{,}362\\
416--515 & 152.8 & 84.6 & 164{,}162\\
\bottomrule
\end{tabular}\end{adjustbox}
\end{sidebody}

%% file: tab/dagger.tex
\begin{sidecap}[0.36]
\caption{\textbf{DAgger labels and data} (R2R-CE val-unseen). All models start from Stage~1 and train for one epoch. TL: trajectory length (m).}
\label{tab:dagger}
\end{sidecap}
\begin{sidebody}[0.60]
\setlength{\tabcolsep}{4.2pt}
\footnotesize
\begin{adjustbox}{max width=\linewidth}\begin{tabular}{@{}l ccccc@{}}
\toprule
Training data & SR & SPL & OS & NE & TL\\
\midrule
Stage 1 & 52.4 & 48.7 & 58.0 & 5.71 & 9.4\\
\midrule
DAgger episodes, executed actions & 52.0 & 42.4 & 70.3 & 5.93 & 16.0\\
DAgger episodes, oracle relabeling & 58.8 & 45.0 & 72.7 & 4.75 & 15.7\\
+ STOP correction & 60.7 & 48.8 & 73.2 & 4.75 & 15.1\\
\rowcolor{oursbg}+ demonstrations (Stage 2) & \textbf{63.7} & \textbf{56.7} & 70.6 & \textbf{4.23} & 11.2\\
\bottomrule
\end{tabular}\end{adjustbox}
\end{sidebody}

%% file: fig/realworld.tex
\newcommand{\rwepisode}[5]{%
\begin{scope}[shift={(#1,#2)}]

  \node[
    anchor=south west,
    inner sep=0pt,
    text width=8.5cm,
    align=flush left,
    font=\sffamily\scriptsize,
    text=inkC
  ] at (0,1.52) {
    \textbf{#3}\enspace\textit{``#4''}
  };

  \foreach \img [count=\k from 0] in {#5}{
    \pgfmathsetmacro{\xx}{1.025+2.1667*\k}

    \node[
      anchor=north,
      inner sep=0pt,
      outer sep=0pt,
      draw=black!22,
      line width=0.35pt
    ] at (\xx,1.413) {
      \includegraphics[width=2.05cm]{\img}
    };
  }

\end{scope}%
}

\resizebox{\linewidth}{!}{%
\begin{tikzpicture}[fig,x=1cm,y=1cm]

\rwepisode{0}{0}{(a) T1 instruction:}
{\textbf{Go straight}, \textbf{pass the foosball table}, continue \textbf{past the pool table}, and \textbf{stop in front of the wooden cabinet.}}
{{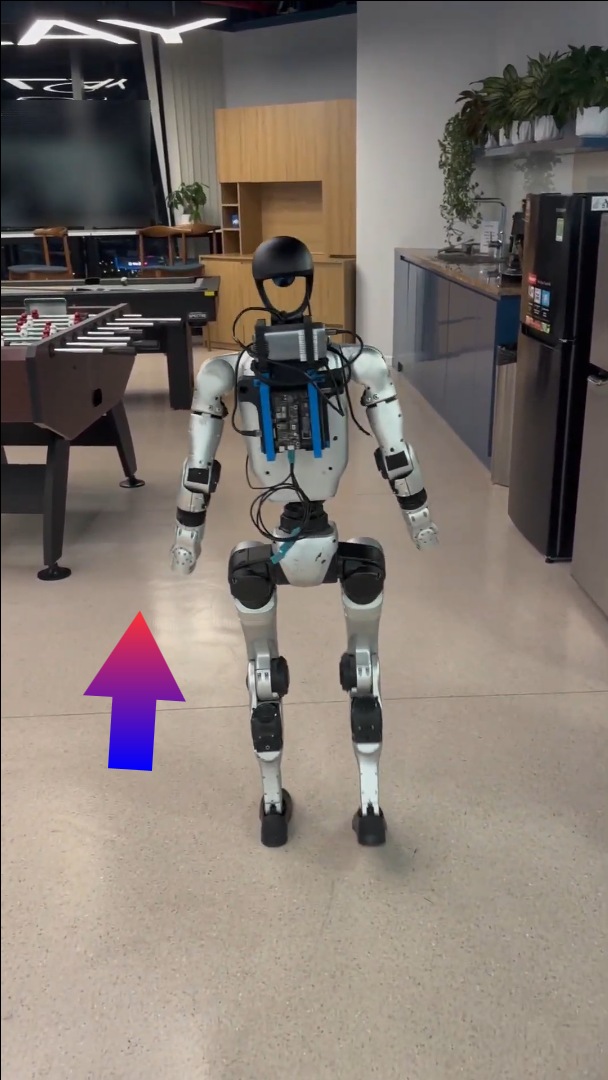},
 {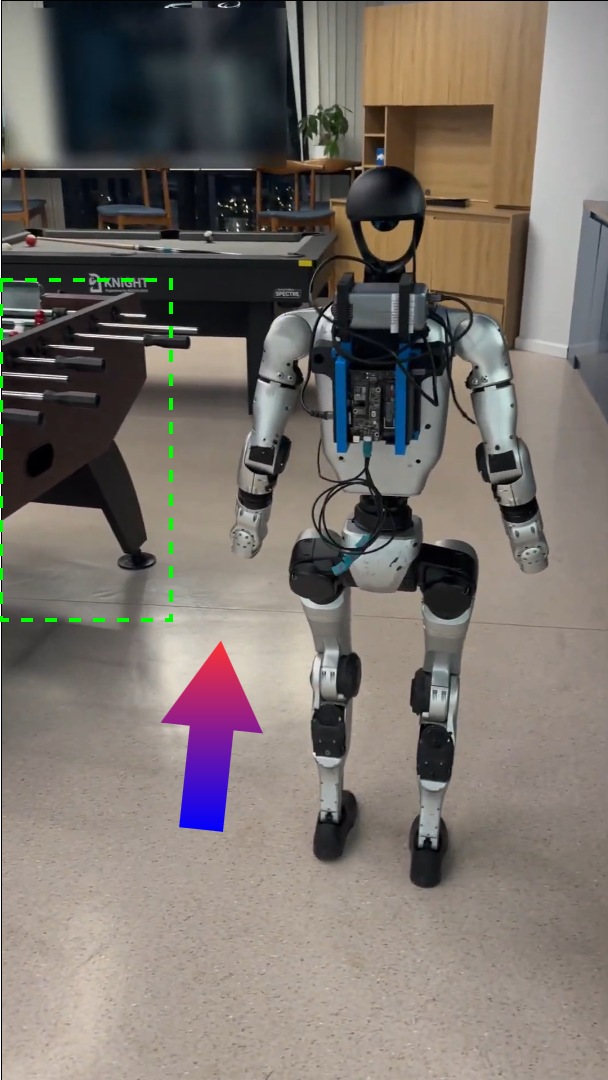},
 {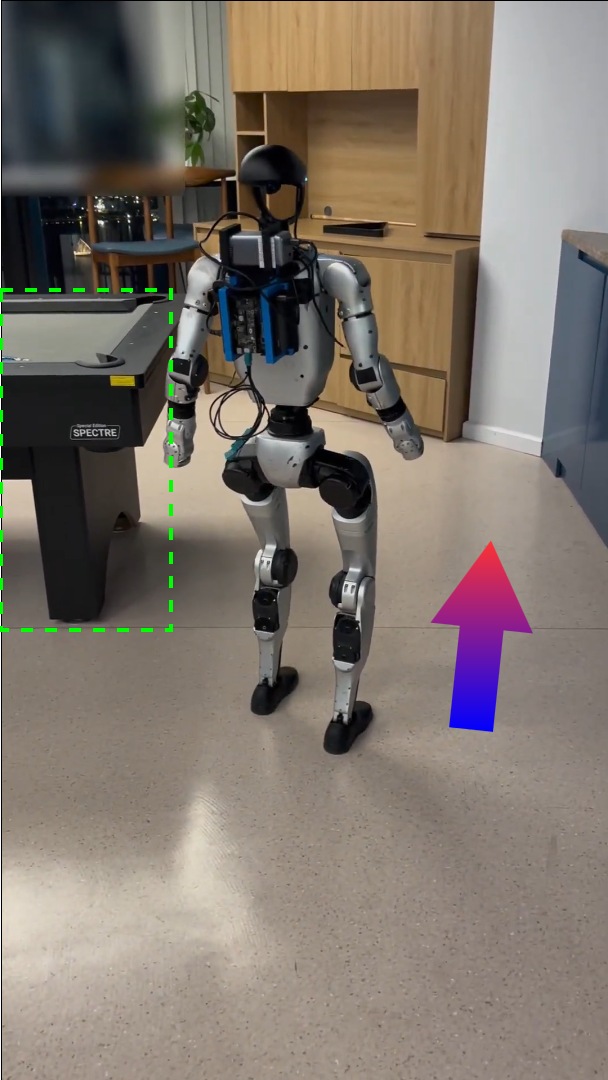},
 {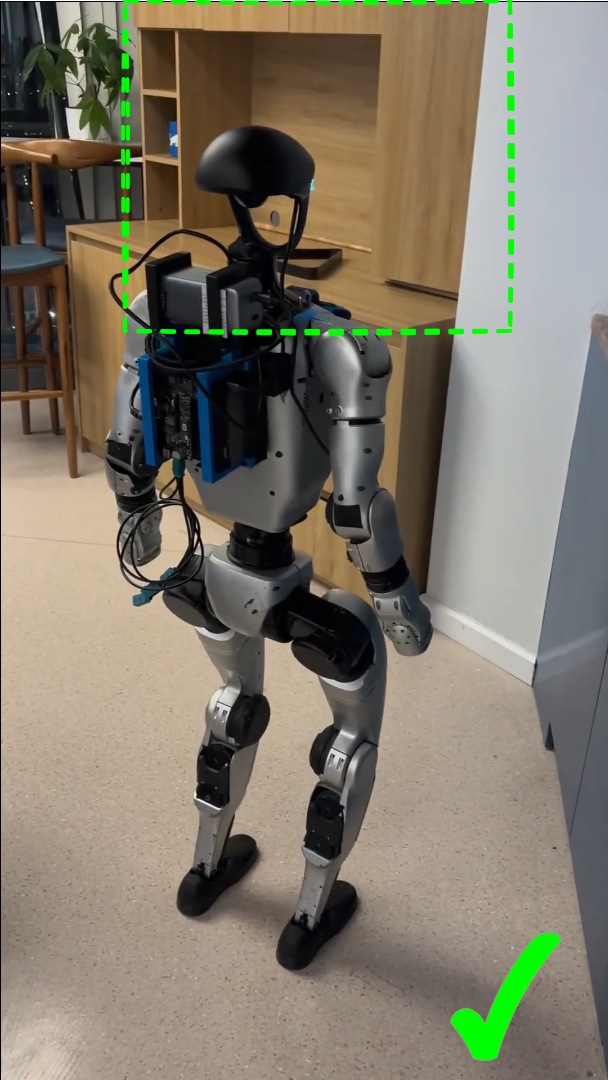}}

\rwepisode{8.91}{0}{(b) T2 instruction:}
{Go straight until you \textbf{reach the red fire extinguisher cabinet},
then \textbf{turn right}. \textbf{Continue straight along the hallway} and \textbf{stop
in front of the potted plant} at the end.}
{{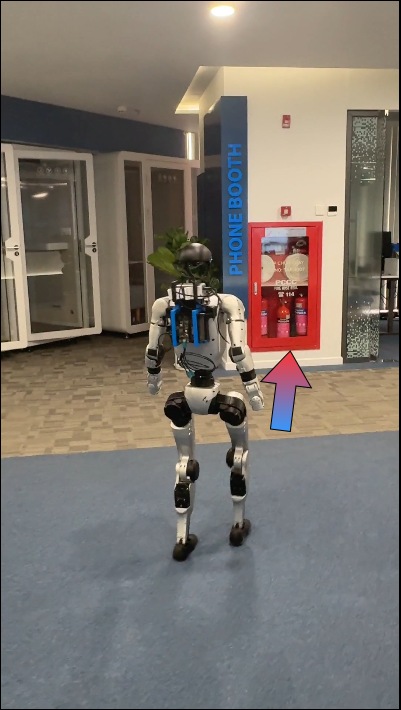},
 {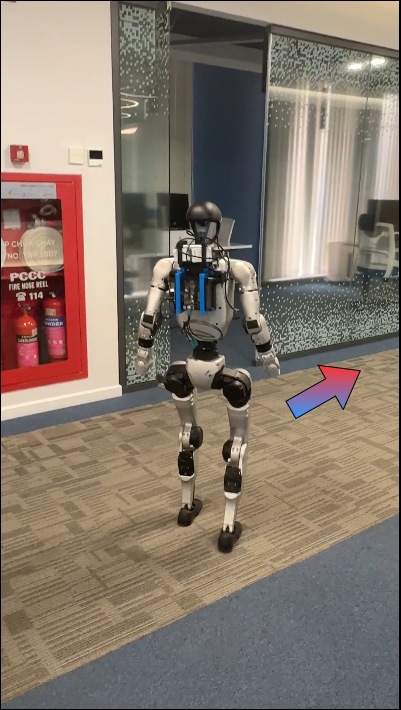},
 {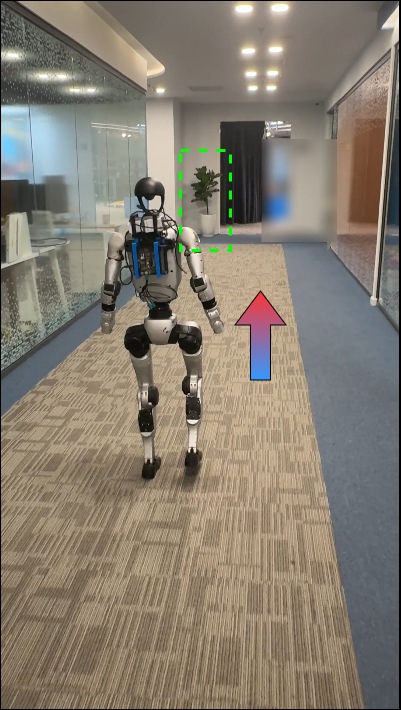},
 {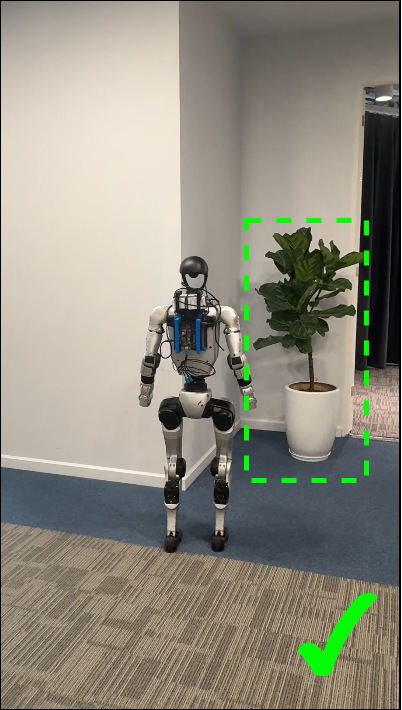}}

\end{tikzpicture}}

\caption{
  \textbf{Real-world deployment of \method} on a Unitree G1 humanoid.
  Each panel shows four key moments during instruction execution,
  ordered from left to right, with the final image showing where the robot stops.
}
\label{fig:realworld}

%% file: sec/5_conclusion.tex
\section{Conclusion}
\label{sec:conclusion}

We presented \method, a VLN agent that encodes each frame once and keeps a constant-size streaming memory: the recurrent state of a hybrid MLLM, a short window of recent observations, and a bank of keyframes that changed the state the most.
Because the memory is causal and bounded, \method is trained on whole episodes with its deployment memory, runs at a constant cost per step, and can copy its full memory in $\mathcal{O}(1)$.
Counterfactual Rewind uses these copies to correct drifting rollouts at their closest approach to the goal and to give counterfactual credit to corrections and continuations.
Using only the R2R-CE and RxR-CE training splits and 1{,}000 RL instructions, a 4B \method reaches 71.9 SR on R2R-CE val-unseen and 73.6 on RxR-CE at 104\,ms per step.

%% file: sec/A_appendix.tex
\section{Additional Real-World Experiments}
\label{sec:supp_real}
\setlength{\intextsep}{7pt plus 2pt minus 2pt}

We provide additional real-world navigation episodes to
complement the experimental results presented in the main
paper. These episodes cover both T1 (moderate-length navigation)
and T2 (long-range navigation) tasks, further illustrating
the ability of \method to follow natural language instructions
and navigate to the specified target locations in real-world
environments.

\subsection{T1: Moderate-length Navigation}
\label{sec:supp_real_t1}

\Cref{fig:supp_real_t1} presents additional T1 navigation
episodes, where the robot must distinguish visually similar
objects and identify the correct stopping location based on
their spatial or sequential relationships.

\begin{figure}[!ht]
\input{fig/realworld_t1}
\end{figure}

\subsection{T2: Long-range Navigation}
\label{sec:supp_real_t2}

\Cref{fig:supp_real_t2} presents additional T2 navigation
episodes involving longer trajectories, landmark recognition,
and multiple navigation decisions along the route.

\begin{figure}[!ht]
\input{fig/realworld_t2}
\end{figure}

%% file: fig/realworld_t1.tex
\noindent \rwpanel[2]{Episode 1.}
{\textbf{Go straight}, \textbf{pass the foosball table}, and \textbf{stop beside the pool table.}}
{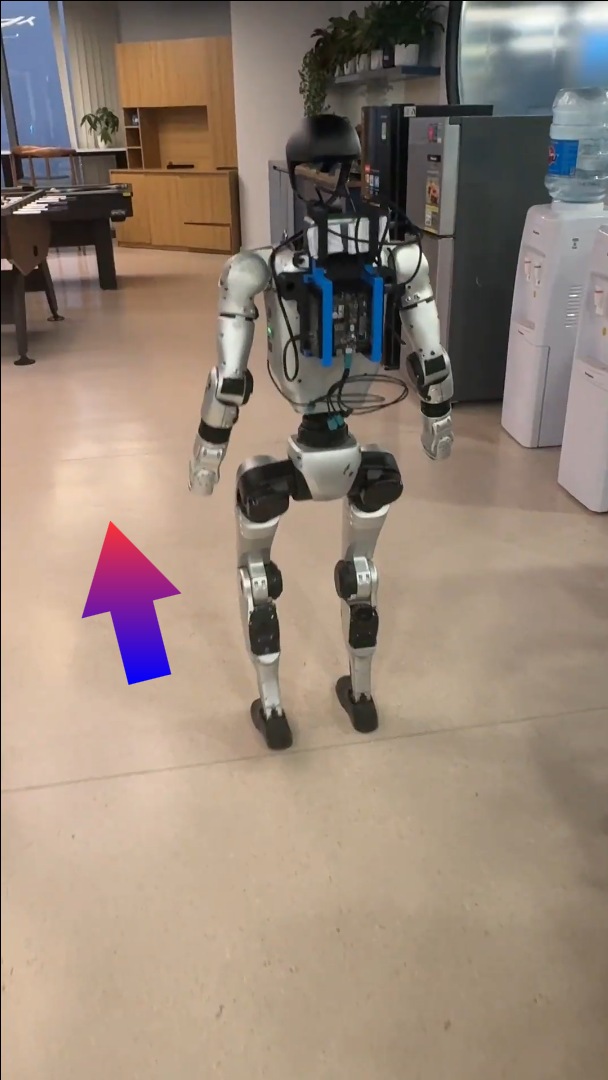}
{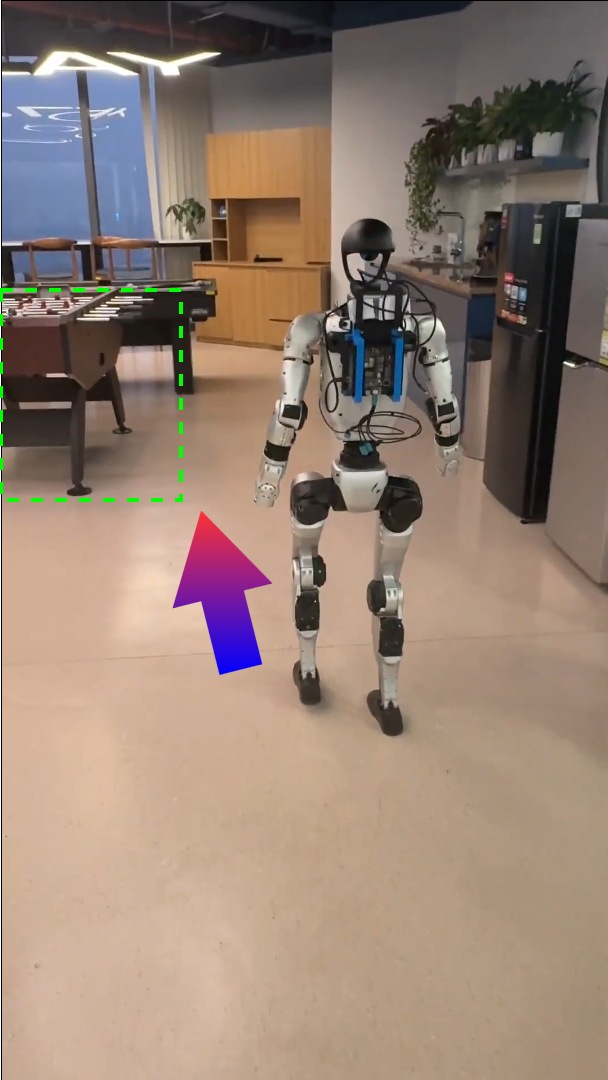}
{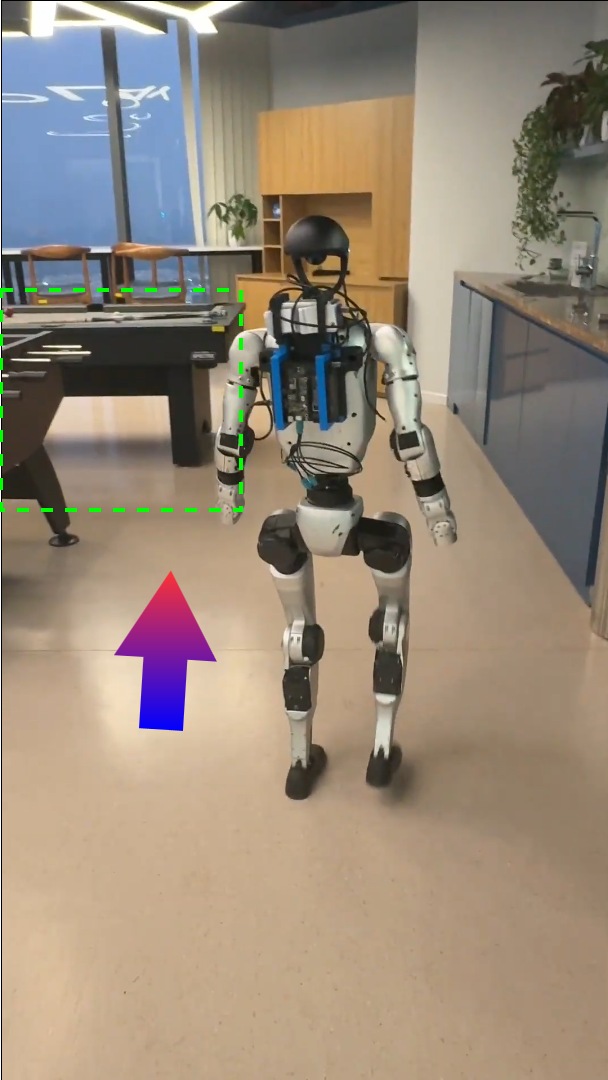}
{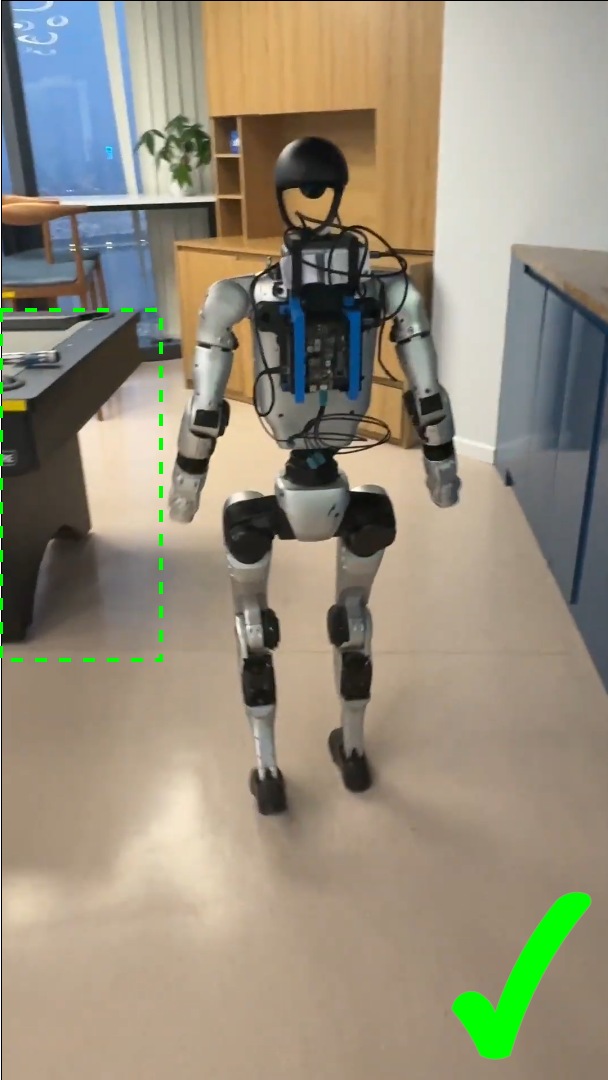}\hfill
\rwpanel[2]{Episode 2.}
{\textbf{Go straight} until you \textbf{reach the red fire extinguisher cabinet}, then \textbf{turn left}, \textbf{continue straight}, and \textbf{stop near the potted plant on your right.}}
{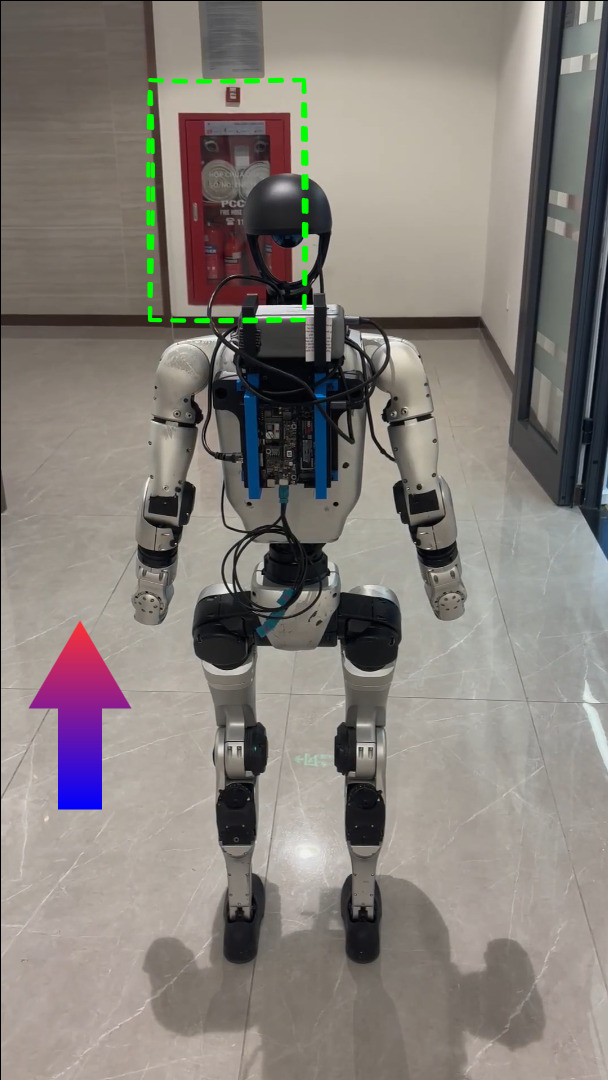}
{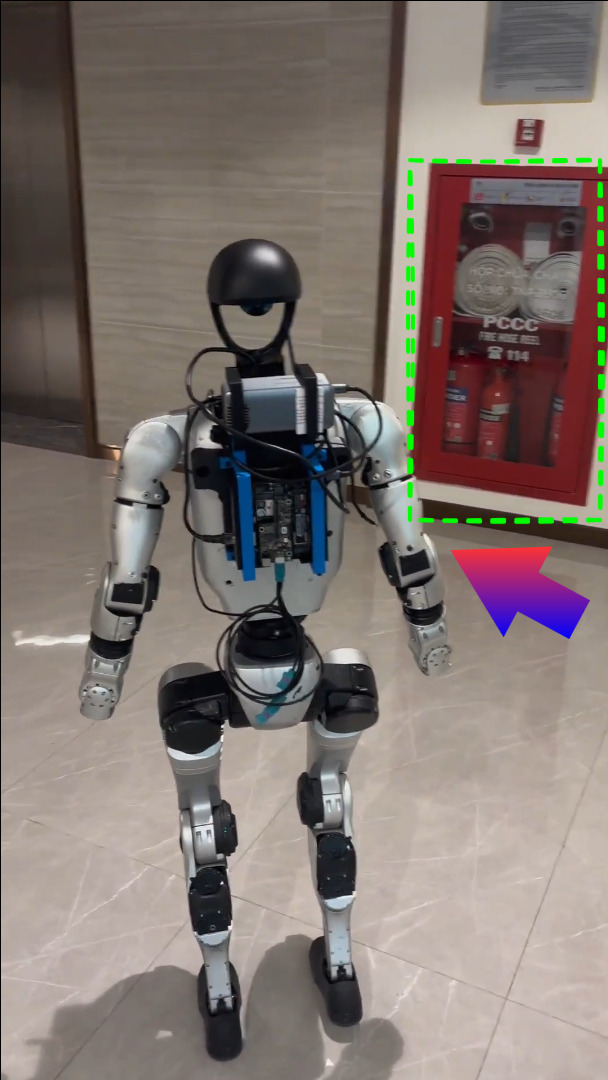}
{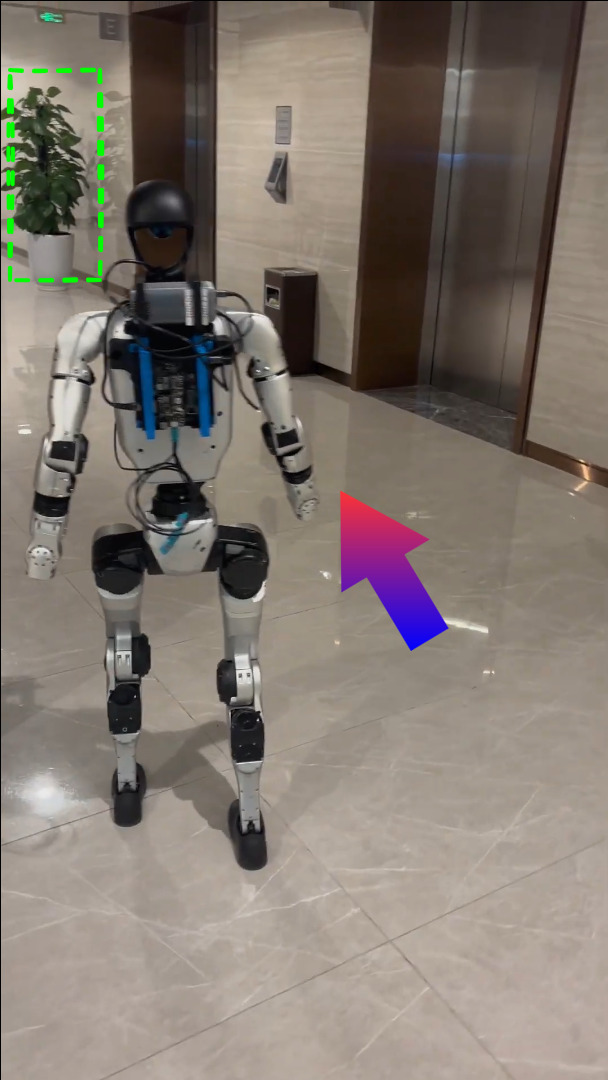}
{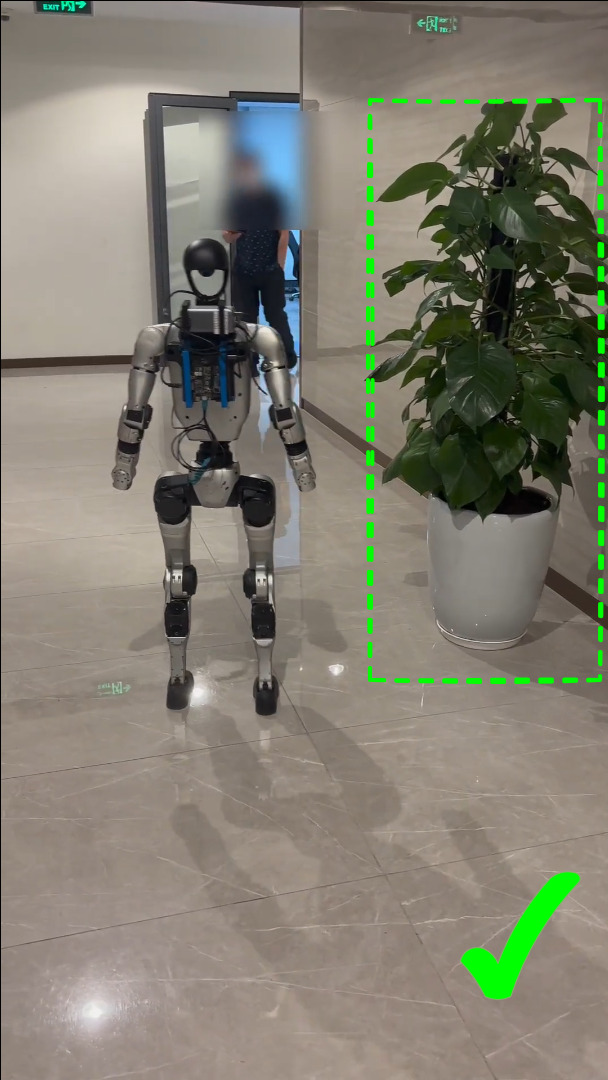}\par
\captionsetup{skip=4pt}\caption{\textbf{Additional real-world navigation episodes} from the T1 (moderate-length navigation) category. Each panel shows four keyframes arranged chronologically from left to right, illustrating the robot's ability to identify the correct target location based on spatial and sequential relationships between objects.}
\label{fig:supp_real_t1}

%% file: fig/realworld_t2.tex
\noindent \rwpanel[2]{Episode 1.}
{\textbf{Go straight into the room ahead}, continue to the \textbf{white table}, then \textbf{turn right}. \textbf{Continue straight} and \textbf{stop in front of the TV.}}
{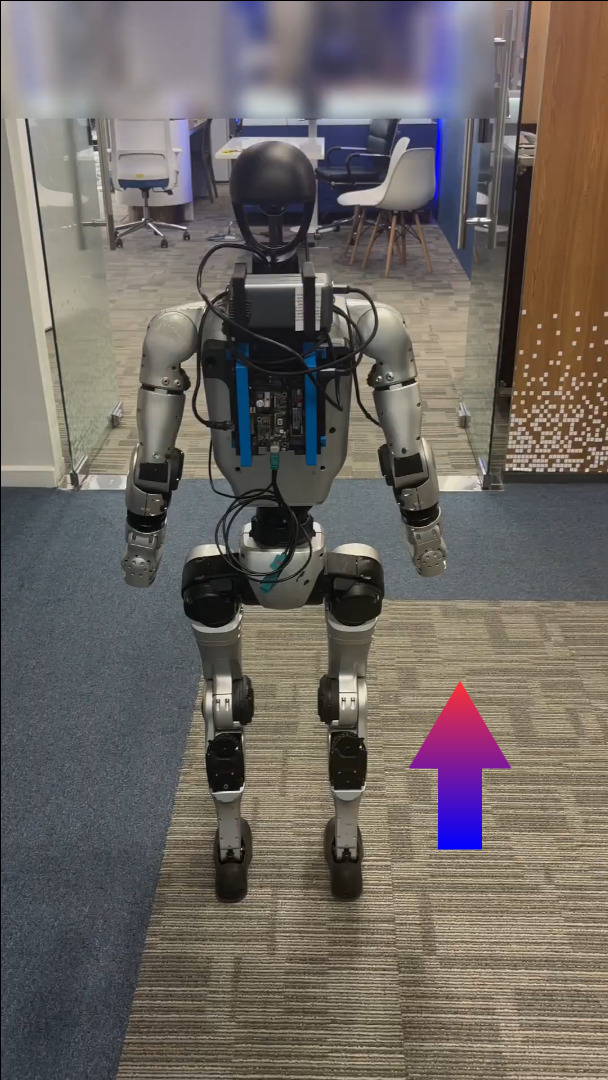}
{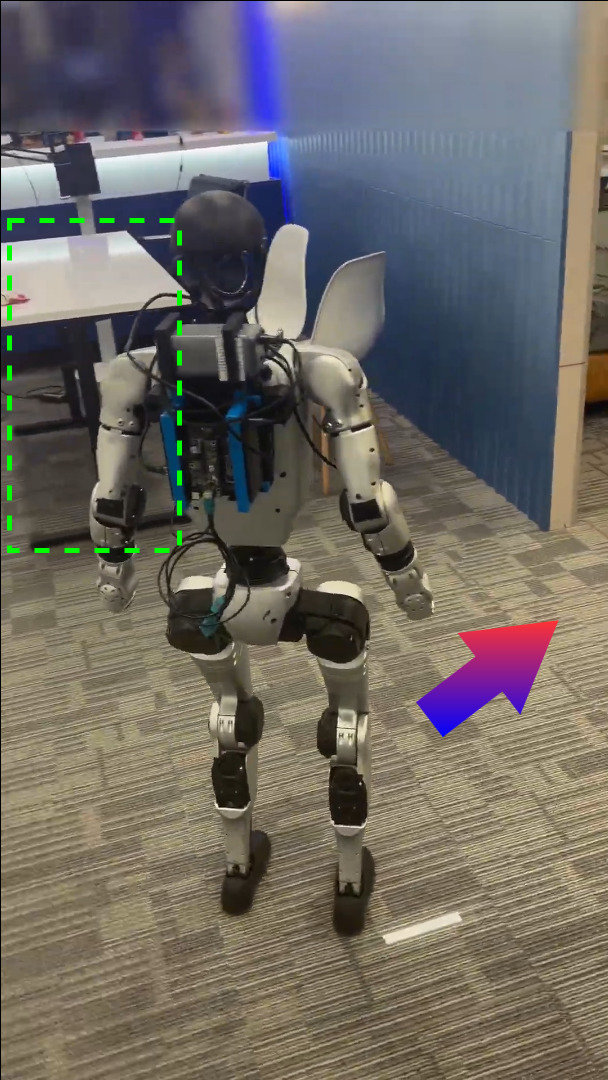}
{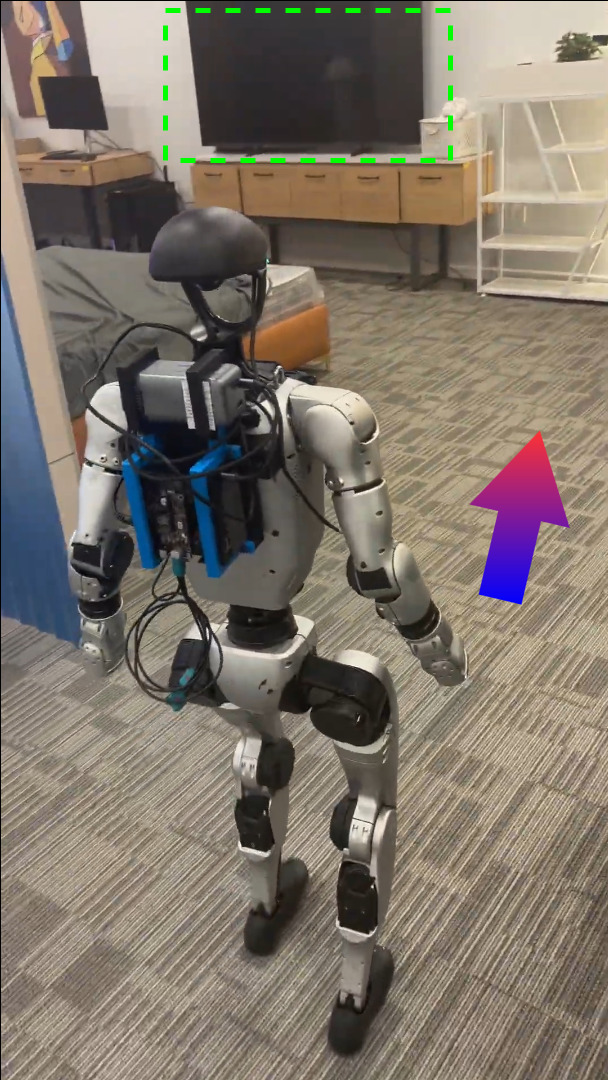}
{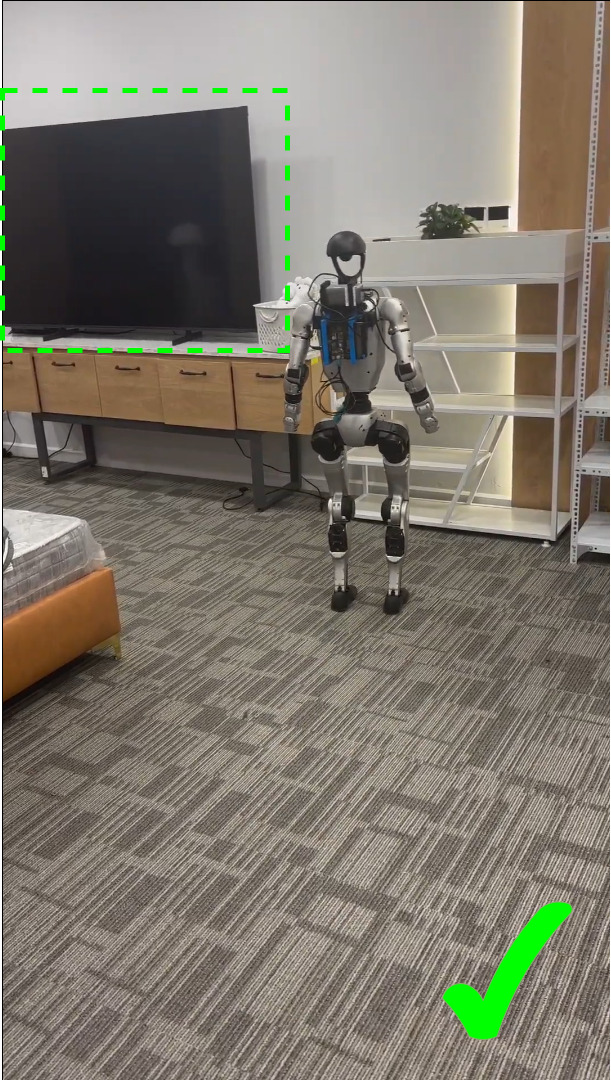}\hfill
\rwpanel[2]{Episode 2.}
{\textbf{Go straight along the hallway}, then \textbf{turn right} at the end to enter the room. \textbf{Continue to the white table} and \textbf{stop} in front of it.}
{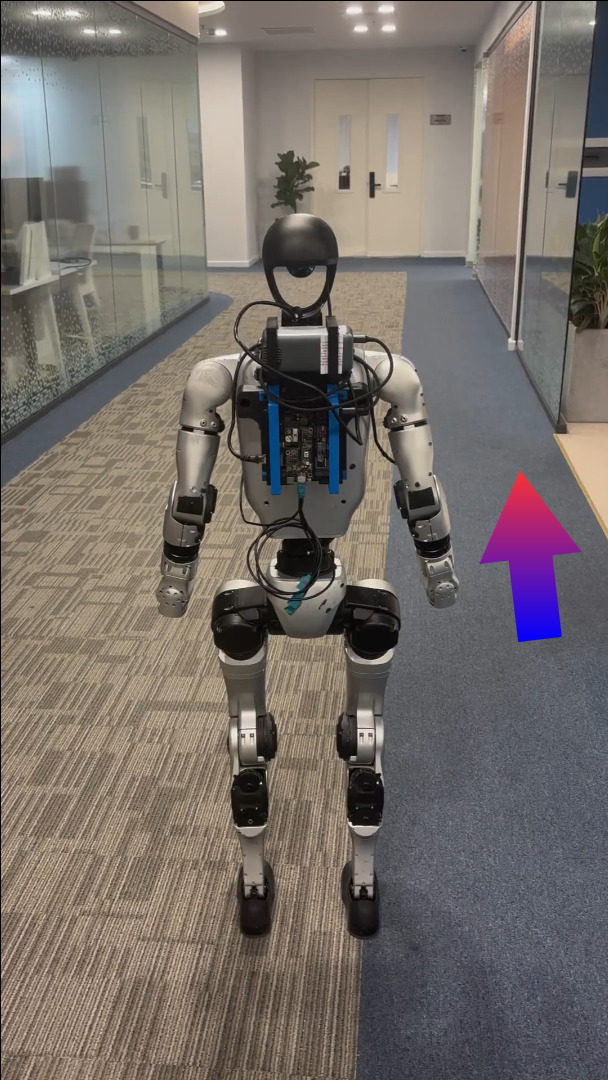}
{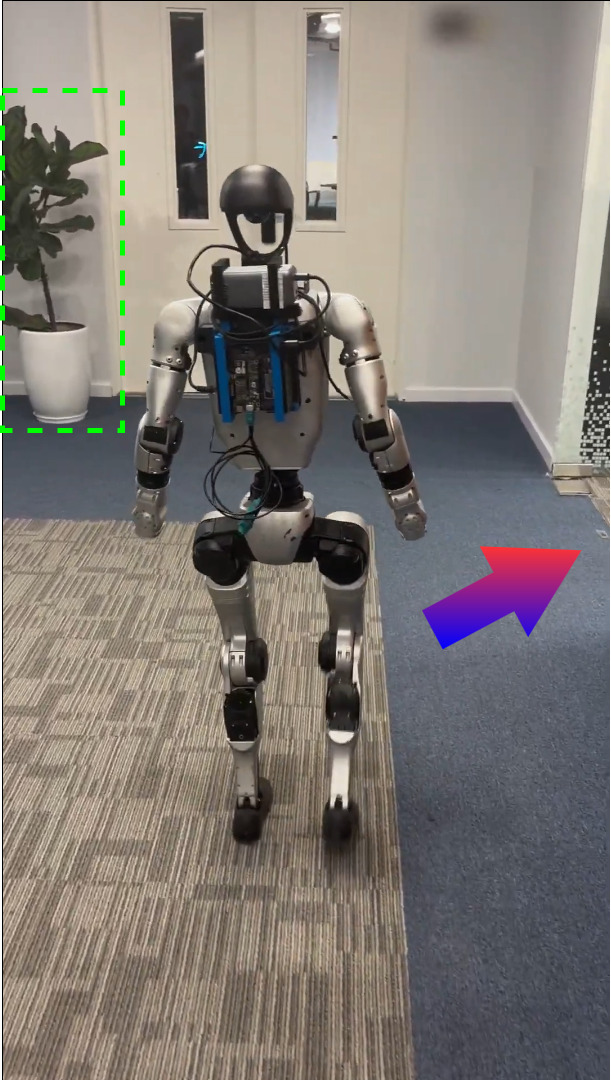}
{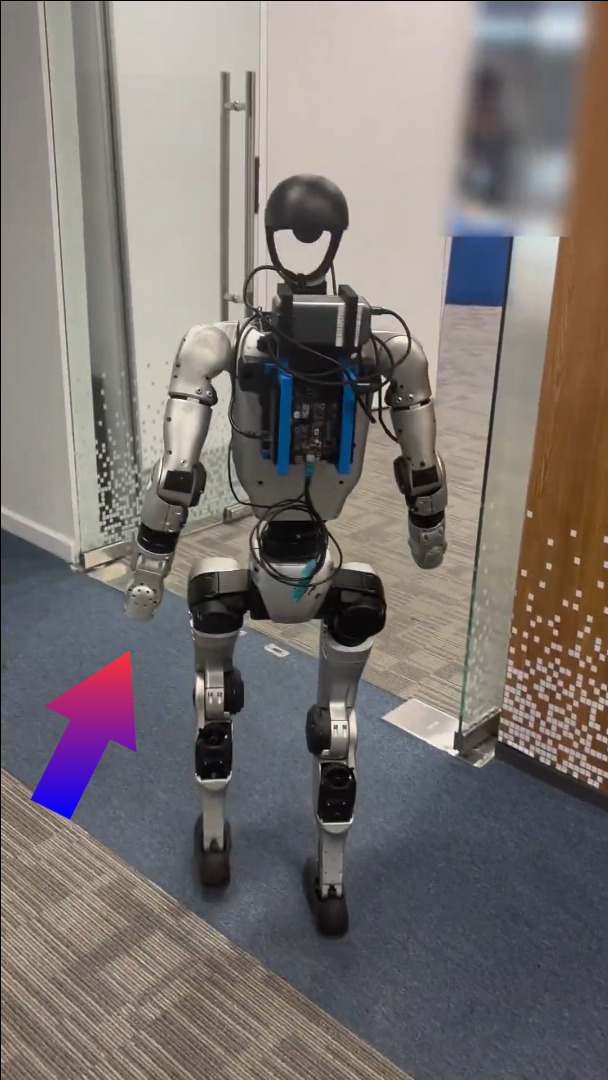}
{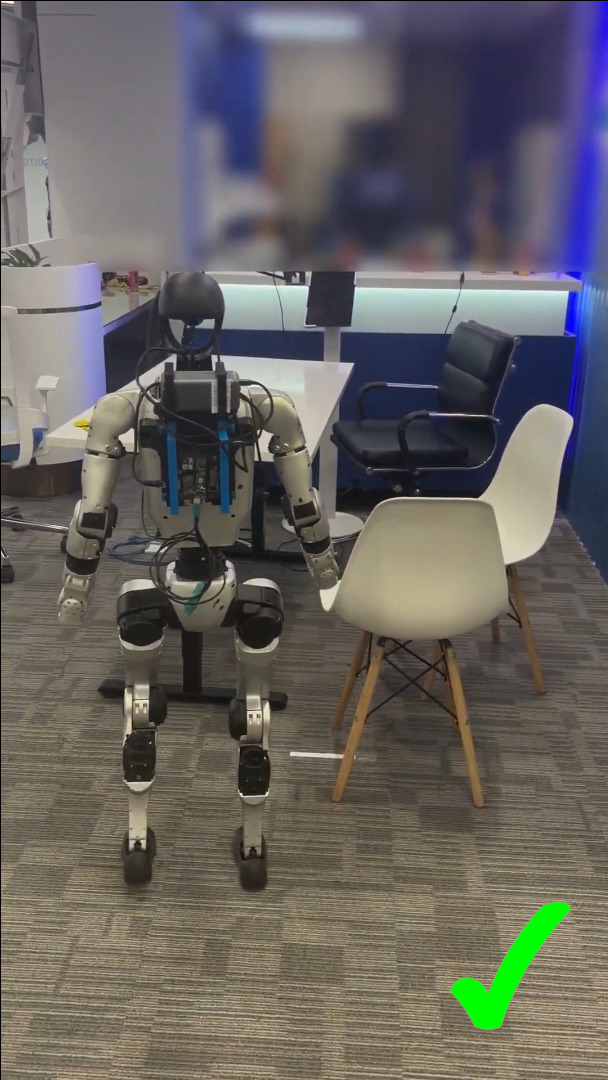}\par
\vspace{5pt}
\noindent \rwpanel[4]{Episode 3.}
{\textbf{Go straight} to the intersection with \textbf{the red fire extinguisher box, turn right}, \textbf{continue straight for a short distance,} then \textbf{turn right into the pantry}. \textbf{Continue straight} and \textbf{stop in front of the refrigerator} on the right side of the room.}
{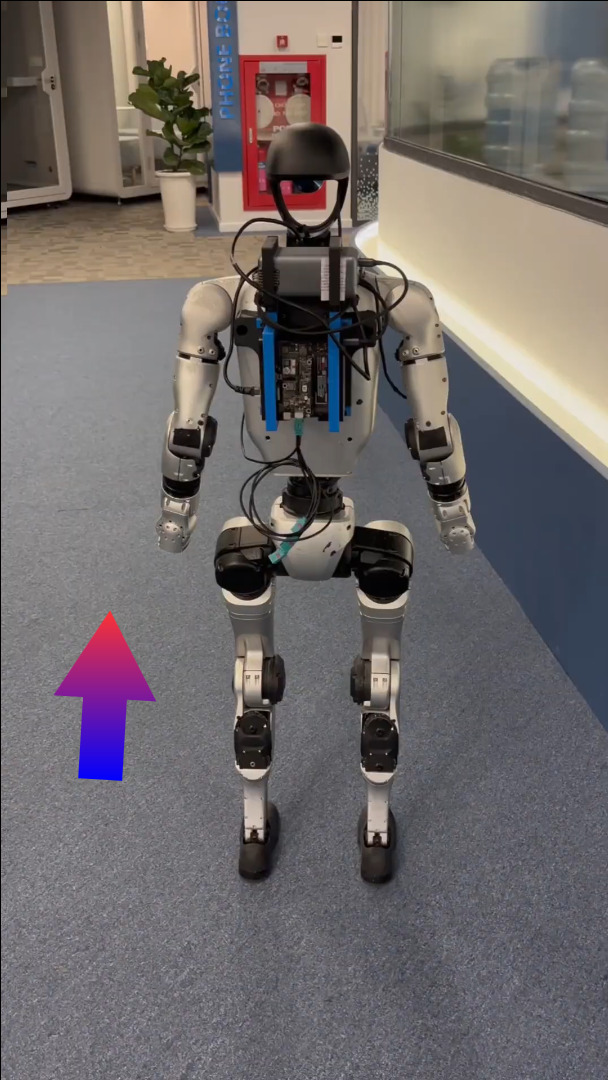}
{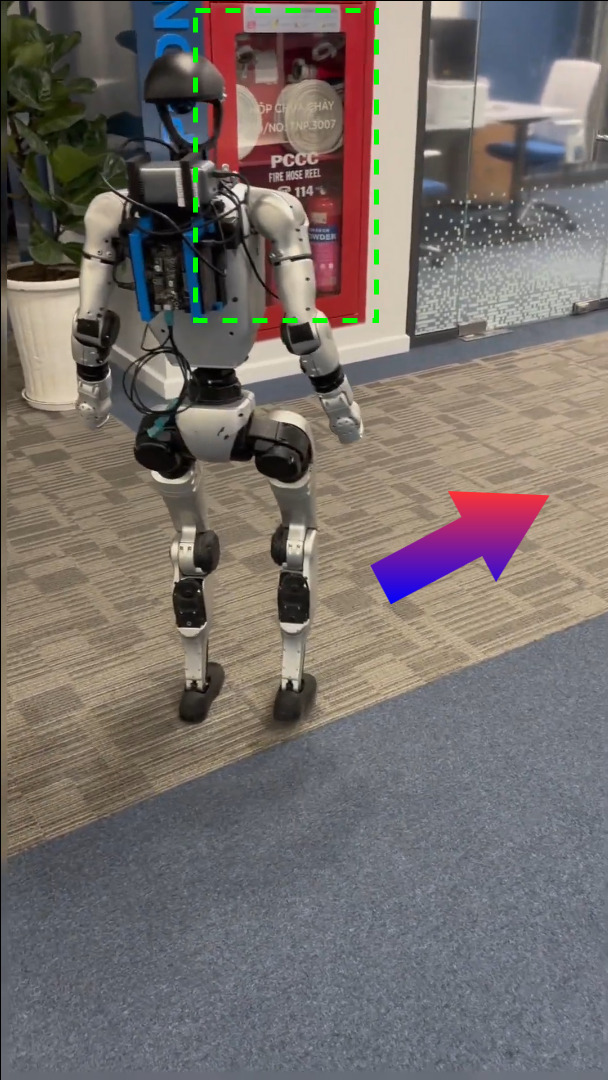}
{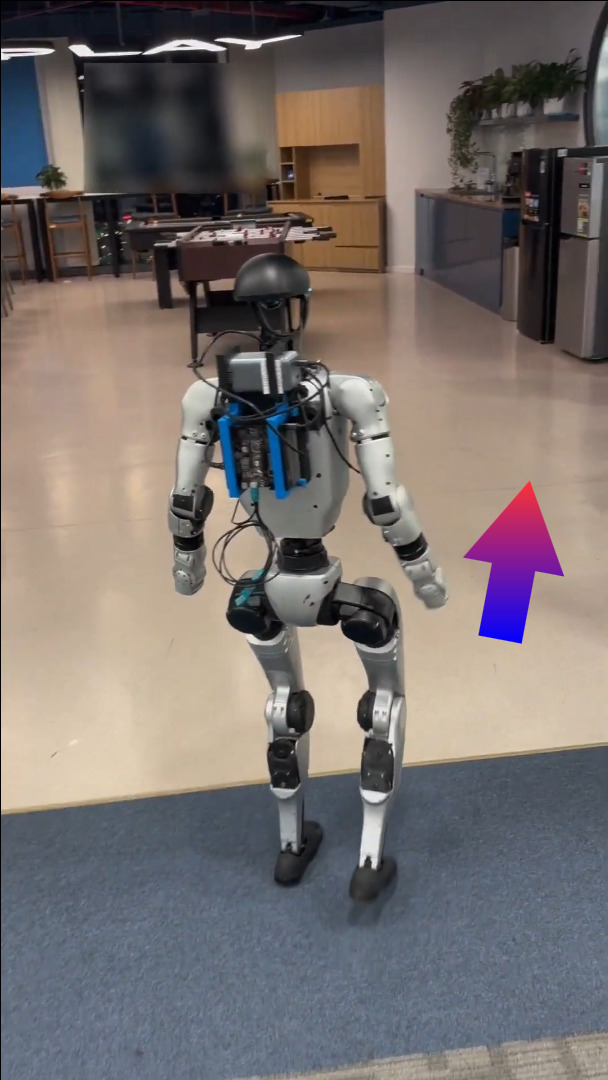}
{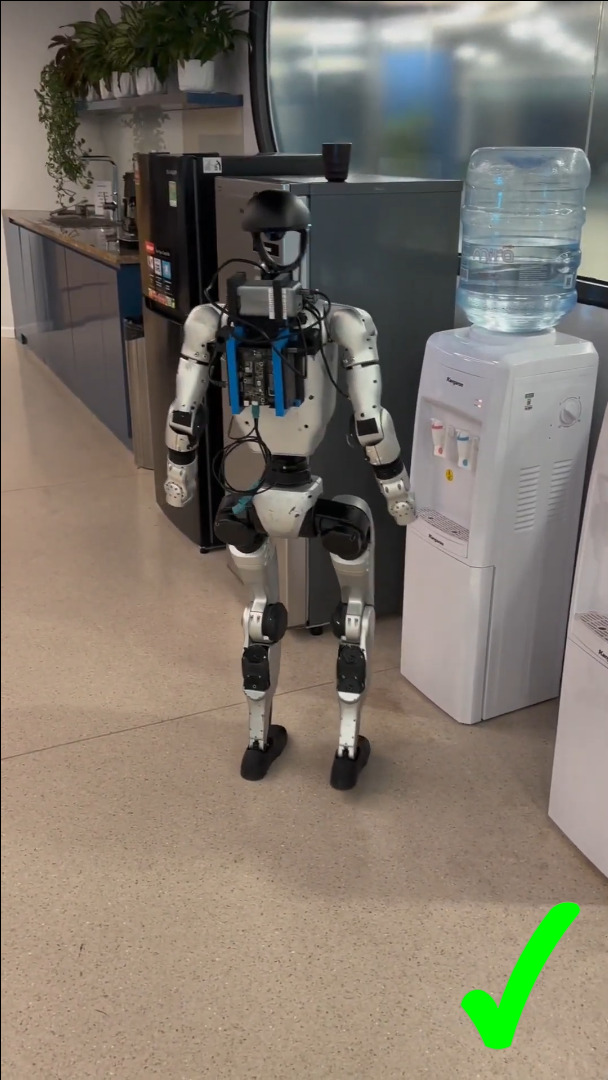}\hfill
\rwpanel[4]{Episode 4.}
{\textbf{Turn left at the intersection}. Continue straight \textbf{past a potted plant} on your right, then keep \textbf{going straight} and \textbf{stop in front of the next potted plant} near the blue couch.}
{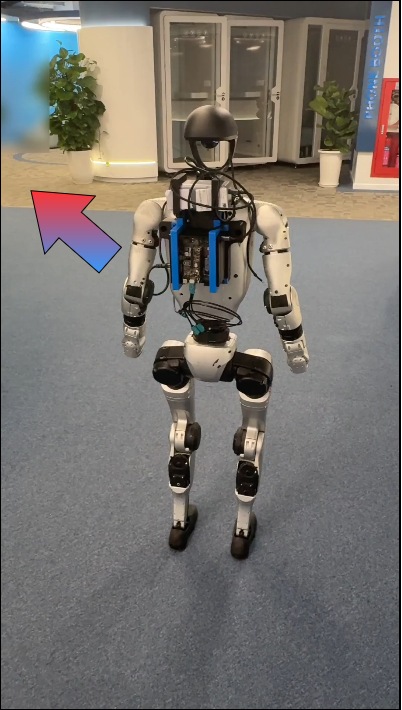}
{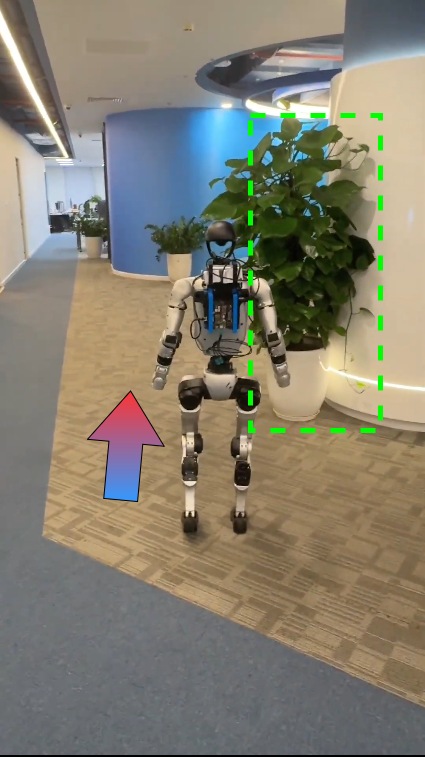}
{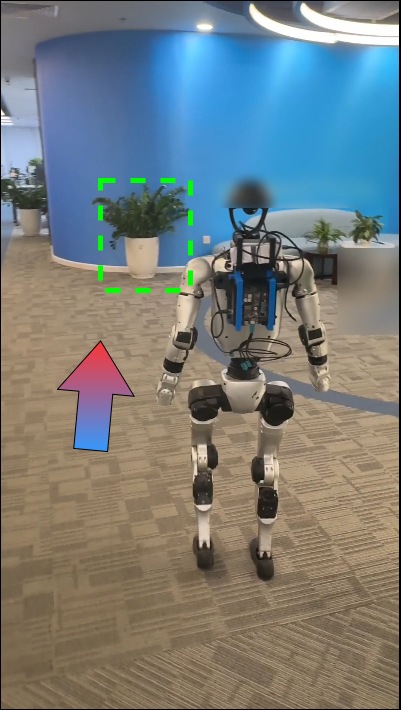}
{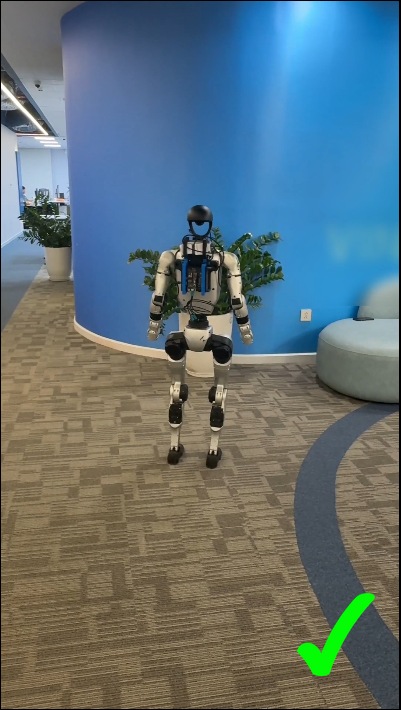}\par
\captionsetup{skip=4pt}\caption{\textbf{Additional real-world navigation episodes} from the T2 (long-range navigation) category. Each panel presents four keyframes arranged chronologically from left to right, showing the robot navigating through multiple decision points and reaching the specified stopping location.}
\label{fig:supp_real_t2}